%% file: main.tex
\documentclass{article}

\usepackage[main,final,nonatbib]{neurips_2026}

\usepackage[utf8]{inputenc} 
\usepackage[T1]{fontenc}    
\usepackage{hyperref}       
\usepackage{url}            
\usepackage{booktabs}       
\usepackage{amsfonts}       
\usepackage{nicefrac}       
\usepackage[patch=none]{microtype}     
\usepackage{xcolor}         
\usepackage{tcolorbox}
\usepackage{multirow} 
\usepackage{array}      
\usepackage{tabularx}
\usepackage{setspace}
\usepackage{amsmath}
\usepackage[scaled=0.96]{helvet}
\usepackage{newtxtext}
\usepackage{sourcesanspro}
\usepackage{listings}
\usepackage{enumitem}
\usepackage{url}
\usepackage{xurl} 

\usepackage{caption}

\usepackage{wrapfig}

\usepackage{makecell}

\usepackage{subcaption}

\usepackage{xspace}

\usepackage{etoc}
\usepackage{titletoc}
\newcommand{\llmicon}[1]{%
    \raisebox{-0.15em}{\includegraphics[height=1em]{llm_icons/#1}}%
    \hspace{0.25em}%
}

\newcommand{\OpenAIIcon}{\llmicon{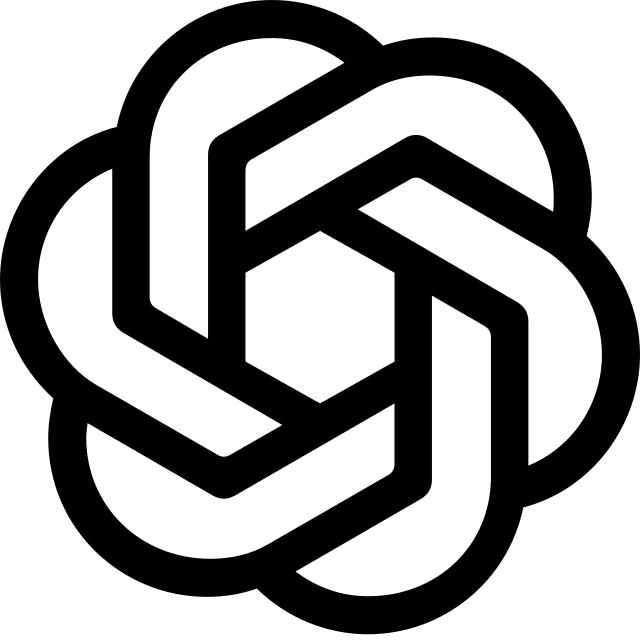}OpenAI\xspace}
\newcommand{\AnthropicIcon}{\llmicon{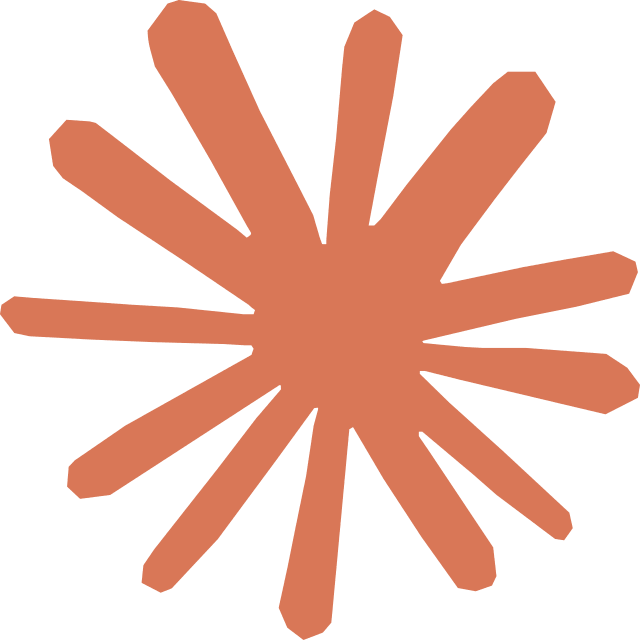}Anthropic\xspace}
\newcommand{\GoogleIcon}{\llmicon{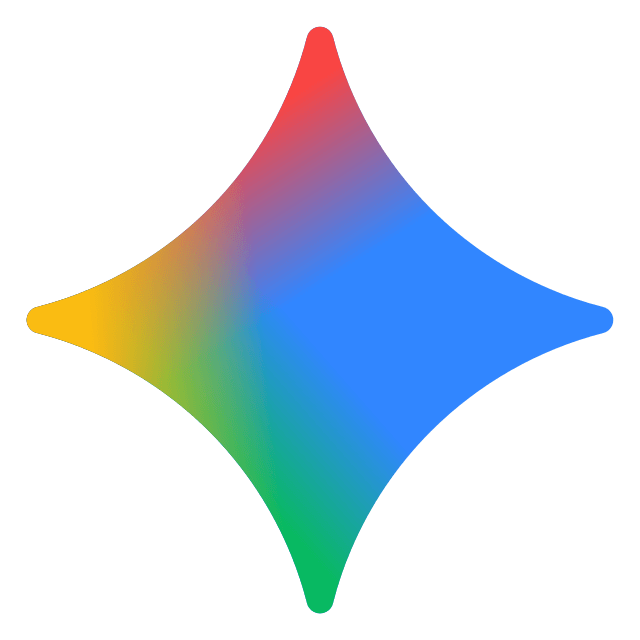}Google\xspace}
\newcommand{\DeepSeekIcon}{\llmicon{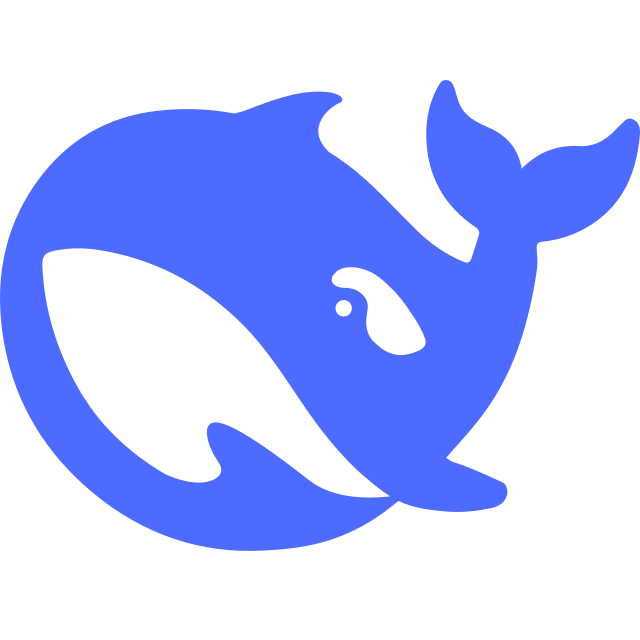}DeepSeek\xspace}
\newcommand{\QwenIcon}{\llmicon{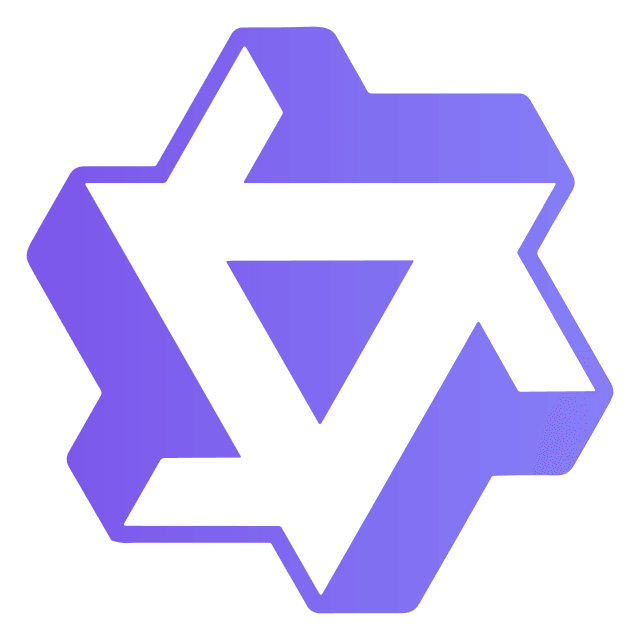}Qwen\xspace}

\definecolor{deepPurple}{RGB}{110, 70, 170}
\definecolor{softPurple}{RGB}{140, 110, 200}
\definecolor{rosePink}{RGB}{210, 90, 140}
\definecolor{softPink}{RGB}{230, 130, 170}
\definecolor{accentBlue}{RGB}{70, 130, 200}

\definecolor{vividPurple}{RGB}{120, 0, 220}
\definecolor{myPurple}{RGB}{102, 51, 153}
\definecolor{myPink}{RGB}{220, 80, 140}

\definecolor{codeblue}{RGB}{0, 0, 180}
\definecolor{codebg}{RGB}{245,245,245}

\newcommand{\thickhline}{\noalign{\hrule height 0.8pt}}

\usepackage{url}

\title{SimpleEvol: An Agent-Loop Framework for LLM-Driven Automated Heuristic Design with Minimal Human Priors}

\author{
Jianghan Zhu$^{1}$ \qquad
Cong Zhang$^{2}$
\qquad
Rongjie Zhu$^{3}$ \qquad
Chi Zhang$^{4}$ \qquad
Zhiguang Cao$^{1}$ \\[4pt]
$^{1}$School of Computing and Information Systems, Singapore Management University, Singapore \\
$^{2}$Independent Researcher \\
$^{3}$School of Teacher Education, Nanjing University of Information Science and Technology, China \\
$^{4}$Department of Mathematics, National University of Singapore, Singapore \\[3pt]
\texttt{zhuj0044@e.ntu.edu.sg} \quad
\texttt{cong.zhang92@gmail.com} \quad
\texttt{rzhu114514@gmail.com} \\
\texttt{czhang24@nus.edu.sg} \quad
\texttt{zgcao@smu.edu.sg}
}

\begin{document}

\maketitle

\begin{abstract}
Large language models (LLMs) have emerged as powerful tools for automated heuristic design (AHD), enabling iterative generation and refinement of heuristics. However, the dominant paradigm embeds LLMs as narrow, fixed components, such as crossover or mutation, within heavily hand‑engineered evolutionary frameworks. We argue this misapprehends LLMs. It treats them as specialized tools rather than general reasoners, constrains them to low‑level operations, and underutilizes their autonomy. Moreover, the extensive human priors in these frameworks violate the bitter lesson principle that general methods scaling with computation surpass hand‑crafted solutions. This raises a key question: \textit{which AHD framework designs best convert stronger LLM capabilities into better heuristics?}
To address this, we propose metrics for LLM-driven AHD framework handcraftedness (AHI) and intelligence conversion efficiency (ICE). Evaluating ten LLMs across three challenging combinatorial optimization problems, we obtain a notable finding that frameworks with fewer human priors consistently yield higher ICE. Based on this finding, we propose SimpleEvol, an agent-loop framework for AHD which removes nearly all human priors and allows the LLM to operate autonomously. SimpleEvol consistently achieves the highest ICE, often by a large margin. Our results challenge the trend toward complex AHD pipelines and point to a lighter and more model‑centric alternative, suggesting that reducing human priors is a more effective strategy to scale up with model intelligence. The source code is available at https://github.com/HenryZhu1029/SimpleEvol-Master.
\end{abstract}

\section{Introduction}
\label{sec:intro}


Combinatorial optimization problems (COPs) underlie many real-world decision-making tasks, including route planning, job scheduling, and resource allocation~\cite{desale2015heuristic}. Since most COPs are NP-hard, practitioners rely on carefully designed heuristics to obtain high-quality solutions within acceptable time budgets~\cite{burke2013hyper}. However, manually crafting effective heuristics is labor-intensive, requires deep domain expertise, and typically yields solutions tailored to specific problem settings~\cite{pillay2018hyperheuristics}.

Automated Heuristic Design (AHD) addresses this bottleneck by treating heuristic construction as a higher-level optimization problem~\cite{qu2020general} known as Hyper-Heuristics~\cite{duflo2019gp,drake2020recent,ye2024reevo}. The recent advent of large language models (LLMs) has substantially expanded the scope of AHD by enabling heuristic generation directly in code space, guided by natural-language reasoning and iterative feedback. 
A growding body of recent work \cite{romera2024mathematical,liu2024evolution,ye2024reevo,dat2025hsevo,zheng2025monte,zhang2026makes,malik2026pyvrp,zhang2026agentic} has established that, within automated heuristic evolution pipelines, LLMs can serve in diverse roles, ranging from code generators to search operators, reflectors, and controllers.
Subsequent work has further extended these frameworks with learning-based reinforcement finetuning~\cite{huang2025calm}, and meta-optimizer discovery through meta-learning~\cite{shi2026generalizable}. 
The growing trend is to pursue better results through the design of more complex evolutionary systems with elaborate human‑designed components for exploration control and population management.

However, a fundamental question remains overlooked and empirically underexplored: \emph{does increasing the structural complexity of an AHD framework actually lead to better utilization of stronger LLMs?} In practice, stronger models are typically treated as drop-in replacements inside existing frameworks, while the question of how much of model intelligence is actually converted into better optimization outcomes remains largely implicit. 
This constitutes a critical gap in the current era of rapidly advancing foundation models~\cite{kaplan2020scaling, hoffmann2022training}. With LLMs rapidly advancing in reasoning, coding, and instruction following, performance gains reported by AHD frameworks increasingly reflect both framework design and model intelligence, making it difficult to judge whether those improvements come mainly from a more effective framework that efficiently leverages those capabilities, or from stronger models themselves. A cross-model view is therefore needed to assess how framework designs scale with model intelligence and whether increasingly elaborate frameworks genuinely help stronger models convert their capabilities into better heuristics.
The bitter lesson~\cite{sutton2019bitter}, a recurring insight from decades of AI research, offers a clarifying perspective that durable progress often comes from systems that lean less on handcrafted structure and more on general-purpose computation. Applied to LLM-based automated heuristic design, this lesson forces a fundamental rethinking. The real question is not which framework yields the best performance on today's benchmarks, but which one is designed to grow more efficiently with advancing model capabilities, to turn LLM intelligence into better solutions, not merely to outsmart static problems with ever more engineering.


In this paper, we take a step toward answering how framework complexity affects the conversion of model intelligence systematically. Our main contributions are as follows:

\begin{itemize}

\item We introduce the \textbf{AHD Handcraftedness Index (AHI)} and \textbf{Intelligence Conversion Efficiency (ICE)} as unified analytical metrics for LLM-based AHD. AHI quantifies the human handcraftedness of an AHD framework along three dimensions: functional complexity, LLM operation diversity, and interaction scale. ICE measures how effectively a framework translates model intelligence gains into downstream heuristic quality, operationalized as the regression slope of framework performance against a unified model intelligence metric across a diverse set of backbone LLMs. Through systematic experiments across 10 backbone LLMs spanning a wide range of intelligence levels and three COPs, we find that \textbf{frameworks with fewer human priors consistently achieve higher ICE}.


\item We propose \textbf{SimpleEvol}, a deliberately minimal agent-loop AHD framework that maximizes model autonomy while minimizing human priors, serving as both a competitive baseline in its own right and a concrete demonstration of our central hypothesis. Through extensive experiments, we find that SimpleEvol achieves the highest ICE on the primary benchmark problems, a finding that remains robust across backbone models, estimator choices, and budget-dependent evaluations. Beyond its higher ICE, SimpleEvol also attains the best performance
under a majority of backbone models on each task, showing that its conversion advantage
is reflected not only in the global scaling trend but also in direct model-wise comparisons. Moreover, SimpleEvol occupies a favorable position on the cost-performance Pareto frontier, demonstrating that its intelligence conversion advantage is not obtained through disproportionate computational expenditure.
\end{itemize}

Our results suggest that as LLMs grow more capable, the primary bottleneck in LLM-based AHD is shifting from framework engineering to model intelligence itself, and that lightweight, model-centric agent-loop frameworks may be better positioned to exploit future advances in foundation models.

\section{Preliminaries}
\label{sec:prelimaries}

\subsection{AHD and LLM-based AHD}
\label{sec:llm-based AHD}
\textbf{AHD.} We formally define the heuristic design problem as an optimization task over a program space $\mathcal{H}$. Let $\mathcal{P}$ denote a combinatorial optimization problem with instance space $\mathcal{I}$ and solution space $\mathcal{S}$. Following prior works \cite{romera2024mathematical,liu2024evolution}, a heuristic $h \in \mathcal{H}$ constructs or improves a feasible solution $h(x) \in \mathcal{S}$ for an instance $x \in \mathcal{I}$.    
Given a task-specific training dataset $D_{\mathrm{train}} \subset \mathcal{I}$ and an objective function $f:\mathcal{S}\rightarrow \mathbb{R}$, the quality of a heuristic is measured by its expected performance on the training distribution. 
For minimization problems, we write
$g(h)=\mathbb{E}_{x\in D_{\mathrm{train}}}\left[-f(h(x))\right]$, so that larger $g(h)$ indicates a better heuristic. Therefore, the goal of AHD is to identify
$h^*=\arg\max_{h\in\mathcal{H}} g(h)$.

\textbf{LLM-based AHD.} In LLM-based AHD, the exploration of the program space is concretely driven by LLMs that generate and refine heuristics. A candidate heuristic $h \in \mathcal{H}$ is produced and evaluated, and the evaluation feedback guides LLMs in subsequent generations. The heuristics designed are typically the core decision functions in a general framework. Examples include designing a constructive heuristic for Traveling Salesman Problem (TSP) that sequentially selects the next city to visit, or an update rule for the scheduling matrix in a guided local search (GLS)-based solver for the Flow Shop Scheduling Problem (FSSP).

The search process can be viewed as operating over structured heuristic states arising from the interaction between a code-generating LLM and a heuristic evaluator. We represent each candidate heuristic state as $\mathbb{S} = (h, \xi)$, where $\xi$ denotes its associated meta information formulated as $\xi = (g(h), \mathcal{T}, \rho, e, t)$,
$\mathcal{T}$ is a natural-language description of algorithmic design logic, $\rho$ captures runtime-related information such as the total execution time, $e$ denotes error messages indicating invalid executions, and $t$ is an identifier tracking the evaluation order of the heuristic during evolution. This formulation reflects that LLM-based AHD is inherently a stateful search process, where the search trajectory can be accessed with compact but informative states, and the summarization is generated conditioned on richer signals other than fitness values alone.

\subsection{Scaling Behavior and the Bitter Lesson}
\label{sec:bitter_lesson}
The bitter lesson proposed by Rich Sutton suggests that general methods that leverage available computation tend to outperform systems relying on handcrafted structures, and that prior knowledge instilled by humans into the agent system, while sometimes yielding short-term gains, often fails to lead to breakthroughs in the long run, a phenomenon repeatedly observed in modern AI systems \cite{fedus2022switch,yousefi2024learning,srivastava2023beyond}. For example, generalist agents demonstrate that a single unified model can handle diverse tasks without task-specific engineering \cite{reed2022generalist}. Similarly, previous work has argued that the reduction of manually engineered structure in favor of learnable components leads to more scalable and effective systems \cite{sinz2019engineering}. 
These observations suggest that in LLM-based AHD, adding human‑designed complexity, such as elaborate population management or hand‑crafted operators, does not reliably translate into better performance. On the contrary, simpler frameworks, which impose less prescriptive search orchestration and leave greater room for the native capabilities of LLMs, may allow improvements in model capability to translate more directly into solution quality.

\section{Intelligence Conversion Model}
\label{sec:intelligence_conversion_model}

In this section, we first quantify the structural complexity of an AHD framework through the AHD Handcraftedness Index (AHI). We then introduce a model-side intelligence metric $I(m)$ based on external benchmarks, together with a performance metric $P(\mathcal{A}, m)$ that captures the quality of the best-found heuristic. Based on these components, we define Intelligence Conversion Efficiency (ICE) to measure how effectively a framework translates model intelligence into optimization performance.

\subsection{AHD Handcraftedness Index (AHI)}
\label{sec:aci}
To analyze LLM-based AHD frameworks from a systems perspective, we introduce the \emph{AHD Handcraftedness Index} (AHI), a descriptive index of the externally prescribed human-crafted
scaffolding imposed around the LLM during the heuristic search. AHI focuses on framework-level structures that organize how the LLM is invoked, conditioned, and routed throughout heuristic search. Our intuition is that such constraints may limit the LLM's reasoning, planning, and generation capabilities, thereby affecting how effectively model intelligence converts into optimization performance. Importantly, AHI measures the amount of human-designed scaffolding around the LLM (e.g., the number of modules, types of LLM operations), not the implementation effort. Low-level computations, such as the reward calculation or reward back-propagation, are not counted separately. In contrast, a search controller that uses these computations to select candidates or route them to subsequent LLM operations is counted as a framework-level orchestration component.
Formally, given an AHD framework $\mathcal{A}$, we define AHI as the sum of three metrics inspired by the classical Halstead Complexity Measures~\cite{hariprasad2017software}:
\begin{equation}
\mathrm{AHI}(\mathcal{A}) =  M +  K +  \log_{10}(1+Q).
\label{eq:aci}
\end{equation}

Each term in Eq.~\ref{eq:aci} captures a unique aspect of handcrafted framework structure. 
Specifically, (1) $M$ measures \emph{functional complexity}: 
defined as the number of separable, stateful orchestration components that independently determine which candidate is selected or routed to a subsequent LLM operation. Examples include generator agents, population-management modules, or evolution controllers (e.g., harmony search). Notably, low-level algorithmic implementation sophistication, as well as components that only retain or format contextual information, are not counted separately. Their influence is reflected only when they alter the types or frequency of LLM operations, which are captured by $K$ and $Q$. (2) $K$ measures \emph{LLM operation diversity}: the number of distinct LLM invocation types with different search roles, conditioning structures, and output contracts, such as crossover, mutation, or reflection (excluding initial bootstrapping). 
(3) $Q$ measures \emph{interaction scale} via the average total LLM calls on a baseline problem across a set of LLMs. A higher $Q$ indicates that each iteration invokes LLM-based operations more frequently, reflecting the cumulative interactions imposed by the LLM-based operators. 
We use $\log_{10}(1+Q)$ to account for the difference in order-of-magnitude between $Q$ and other terms, preventing it from dominating the overall AHI score. $Q$ naturally increases with the structural complexity of the frameworks (e.g., more intermediate prompting can increase the number of LLM calls per training cycle).



\begin{wraptable}{r}{0.5\textwidth}
\vspace{-10pt}
\centering
\small
\caption{AHI of different AHD frameworks.}
\label{tab:aci_comparison}
\setlength{\tabcolsep}{2.5pt}
\renewcommand{\arraystretch}{1}
\begin{tabular}{lcccc}
\toprule
Framework & $M$ & $K$ & $\log_{10}(1+Q)$ & AHI \\
\midrule
FunSearch   & 3 & 1 & 2.915 & 6.915 \\
EoH         & 2 & 4 & 2.919 & 8.919 \\
ReEvo       & 2 & 4 & 3.112 & 9.112 \\
SimpleEvol (ours)  & 1 & 2 & 3.001 & 6.001 \\
\bottomrule
\end{tabular}
\vspace{-10pt}
\end{wraptable}
\paragraph{Case study.}
Consider EoH~\cite{liu2026eoh} as an example. 
Its indispensable modules include a generator agent (i.e., a single LLM) and a population selection module, giving $M=2$. 
Its LLM operations consist of four evolutionary operators (E1, E2, M1, M2), giving $K=4$. 
Lastly, under the default configuration of EoH, the average total number of LLM calls on TSP under step-by-step constructive framework is $Q=828$, corresponding to $\log_{10}(1+Q)=2.919$. 
Therefore, $\mathrm{AHI}(\text{EoH}) = 2 + 4  + 2.919 = 8.919.$ Herein, AHI provides a unified quantitative view of framework complexity of different AHD systems. 
This enables us to move beyond qualitative descriptions such as ``simple'' or ``complex'', and to systematically study how framework complexity relates to intelligence conversion efficiency in the following section. The resulting AHI values for representative frameworks are summarized in Table~\ref{tab:aci_comparison}. Detailed computation procedures of AHI are provided in Appendix \ref{sec:details_of_AHI}.


\subsection{Model Intelligence Metric}
\label{sec:details_of_i(m)}

To quantify intelligence conversion efficiency,  we first introduce a model-side intelligence metric $I(m)$, where $m$ denotes the backbone LLM. The purpose of $I(m)$ is to provide a framework-agnostic estimate of model intelligence available to an AHD system. Since ICE measures how effectively an AHD framework converts model intelligence into optimization performance, $I(m)$ should capture the intrinsic capabilities of the underlying LLM in knowledge utilization, mathematical reasoning, coding, and instruction following, rather than the outcome of a particular heuristic search process or prompt-engineering strategy. We therefore select four famous benchmarks to cover fundamental LLM capability dimensions closely related to LLM-based AHD to provide a multifaceted yet relevant proxy for the LLM capabilities most likely to influence heuristic generation, debugging, and refinement. Let $\mathcal{B}$ denote the set of external benchmarks for estimating intelligence.
Specifically, $I(m)$ is instantiated using four representative benchmarks widely adopted in modern LLM performance evaluation: 

\begin{equation}
\mathcal{B} = \{\text{MMLU-Pro}~\cite{wang2024mmlu}, \text{IFBench}~\cite{pyatkin2025ifbench}, 
\text{AIME2025}~\cite{ye2025aime}, \text{LiveCodeBench}~\cite{jain2024livecodebench}\}.
\label{eq:llm_benchmark}
\end{equation}

Let $s_{m,b}$ denote the raw performance score of model $m$ on benchmark $b \in \mathcal{B}$. 
As the benchmarks operate on different numerical scales, raw scores are not directly comparable. We thus normalize each benchmark by the score of the weakest model $m'$ (the one with the lowest average on $\mathcal{B}$) and take the geometric mean of the normalized values:
\begin{equation}
I(m) = \left( \prod_{b \in \mathcal{B}} \frac{s_{m,b}}{ s_{m',b}} \right)^{\frac{1}{|\mathcal{B}|}}.
\label{eq:model_intelligence}
\end{equation}

We adopt the geometric mean to preserve relative performance differences between heterogeneous benchmarks while ensuring that no single benchmark disproportionately influences the overall score~\cite{john2006aggregating,mariani2022aggregating}. This aggregation rewards models with balanced and complementary capabilities, which is desirable in LLM-based AHD: effective heuristic generation, refinement, and debugging require not only strong reasoning and coding ability, but also robust instruction following and broad knowledge utilization. The geometric mean thus provides a principled, framework-agnostic indicator of the intelligence available for conversion into downstream optimization performance. 
Detailed benchmark sources and calculations for $I(m)$ are provided in Appendix~\ref{sec:details_I(m)}.

\subsection{AHD Performance Metric}
\label{sec:details_of_p(m)}

Having defined the model-side intelligence measure, we next specify the framework-side performance metric. This metric should capture the final optimization quality achieved by an AHD system under a fixed LLM, while remaining independent of the LLM's benchmark scores. This ensures that ICE reflects a clean conversion between two separate components.
Specifically, let $g_{\mathcal{A},m,r,n}$ denote the mean test gap obtained by the best heuristic discovered by AHD framework $\mathcal{A}$ in run $r$ with model $m$ on problem size $n$. Let $\mathcal{D}^{\mathrm{test}}_{n}$ denote the test set of problem size $n$. The formal definition of $g_{\mathcal{A},m,r,n}$ is: $
g_{\mathcal{A},m,r,n}
=
\frac{1}{|\mathcal{D}^{\mathrm{test}}_{n}|}
\sum_{x\in\mathcal{D}^{\mathrm{test}}_{n}}
\frac{J(\mathcal{A},m,r,x)-J^*(x)}{J^*(x)},
$
where $J(\mathcal{A},m,r,x)$ denotes the objective value achieved on test instance $x$ and $J^*$ denotes the reference objective value (e.g., the best-known or optimal objective score). To reduce run-level variance, we average the test performance over multiple independent runs. Let $R$ be the number of independent runs, and $N$ be the set of evaluated test problem sizes. We first compute the average overall test gap $\overline{\mathrm{g}}$ to reflect the generalization ability of heuristics commonly adopted in MCDM~\cite{triantaphyllou2000multi}: 
\begin{equation}
\overline{\mathrm{g}}(\mathcal{A},m)=
\frac{1}{R|N|}
\sum_{r=1}^{R}\sum_{n}^{N}g_{\mathcal{A},m,r,n}.
\label{eq:avg_gap}
\end{equation}

The overall AHD performance metric is therefore formulated as:
\begin{equation}
P(\mathcal{A},m)=\frac{1}{\overline{\mathrm{g}}(\mathcal{A},m)+\epsilon},
\label{eq:Pm_main}
\end{equation}

where $\epsilon>0$ is a small constant for numerical stability. A larger value of $P$ indicates that framework $\mathcal{A}$ is able to achieve better optimization performance under model $m$. 
We adopt the reciprocal form for two reasons. First, the raw gap is a loss-type quantity, where lower values are preferable, whereas ICE is a utility-type measure that should increase with performance, so the reciprocal transformation accomplishes this inversion. Second, the reciprocal increases sensitivity to improvements in the low-gap regime, where progress is inherently difficult. A framework that achieves gains under such conditions merits a strong positive signal, which the reciprocal provides.
Using Eq.~\ref{eq:Pm_main} thus aligns the metric with the practical objective of heuristic design, namely, distinguishing frameworks by their ability to achieve high-quality heuristics rather than merely coarse improvements.

\subsection{Intelligence Conversion Efficiency (ICE)}
\label{sec:ice_definition}

\textbf{What is ICE}? We introduce \emph{Intelligence Conversion Efficiency} (ICE) to quantify how effectively model intelligence translates into optimization performance. For an AHD framework $\mathcal{A}$, we summarize how $P(\mathcal{A},m)$ varies with $I(m)$ over the evaluated model range in the form of a regression line as
\begin{equation}
P(\mathcal{A},m) \approx \alpha_\mathcal{A} I(m) + \beta_\mathcal{A},
\label{eq:ice_regression}
\end{equation}
where $\alpha_\mathcal{A}$ and $\beta_\mathcal{A}$ are framework-dependent regression coefficients. 
We then define the ICE of framework $\mathcal{A}$ as the regression slope:
\begin{equation}
\mathrm{ICE}(\mathcal{A}) = \alpha_\mathcal{A}.
\label{eq:ice_def}
\end{equation}

This definition has a natural interpretation: $\mathrm{ICE}(\mathcal{A})$ measures the fitted first-order change in framework performance per unit increase in model intelligence. 
A larger ICE indicates that the framework is more effective in translating stronger backbone intelligence into better heuristics. 
\textbf{Why We Use ICE}? 
The key motivation of ICE is to characterize how effectively a framework converts model intelligence into optimization performance empirically in a global view with different levels of intelligence. The regression-based definition naturally provides such a summary by leveraging all evaluated models, yielding a stable estimate that is not tied to any particular reference point. This regression-based view is also in line with empirical scaling analyses, where fitted trends are used to capture how performance changes as model intelligence or scale increases. Although the true relationship between $I(m)$ and $P(\mathcal{A},m)$ may not be strictly linear, the regression slope serves as a robust first-order summary of the conversion trend across the evaluated intelligence scope, consistent with theoretical scaling analyses~\cite{hoffmann2022training,hernandez2021scaling} and empirical evidence showing that model capability correlates approximately linearly with downstream performance~\cite{huang2024compression}.


\textbf{How We Use ICE?} We report regression ICE as the primary metric in our experiments. We also verify the sensitivity of ICE to the benchmark choice in $\mathcal{B}$, evaluation budget $T$, or to individual models $m$ in Appendix~\ref{sec:ICE_variants} and \ref{sec:budget_ice}. Together with AHI, ICE provides the analytical basis for examining whether simpler, less handcrafted frameworks better convert model intelligence into heuristic quality.

\section{The Simple Evolution (SimpleEvol) Framework}
\label{sec:simpleEvol_main}
We introduce \textbf{SimpleEvol}, a deliberately minimal AHD framework designed to maximize model autonomy while minimizing handcrafted constraints. It serves as a competitive AHD method in its own right, and more importantly, as a concrete instantiation to verify our central hypothesis that frameworks aligned with the bitter lesson should yield stronger intelligence conversion efficiency.

\begin{wrapfigure}{r}{0.51\textwidth}
    \vspace{-8pt}
    \centering
    \includegraphics[width=0.51\textwidth]{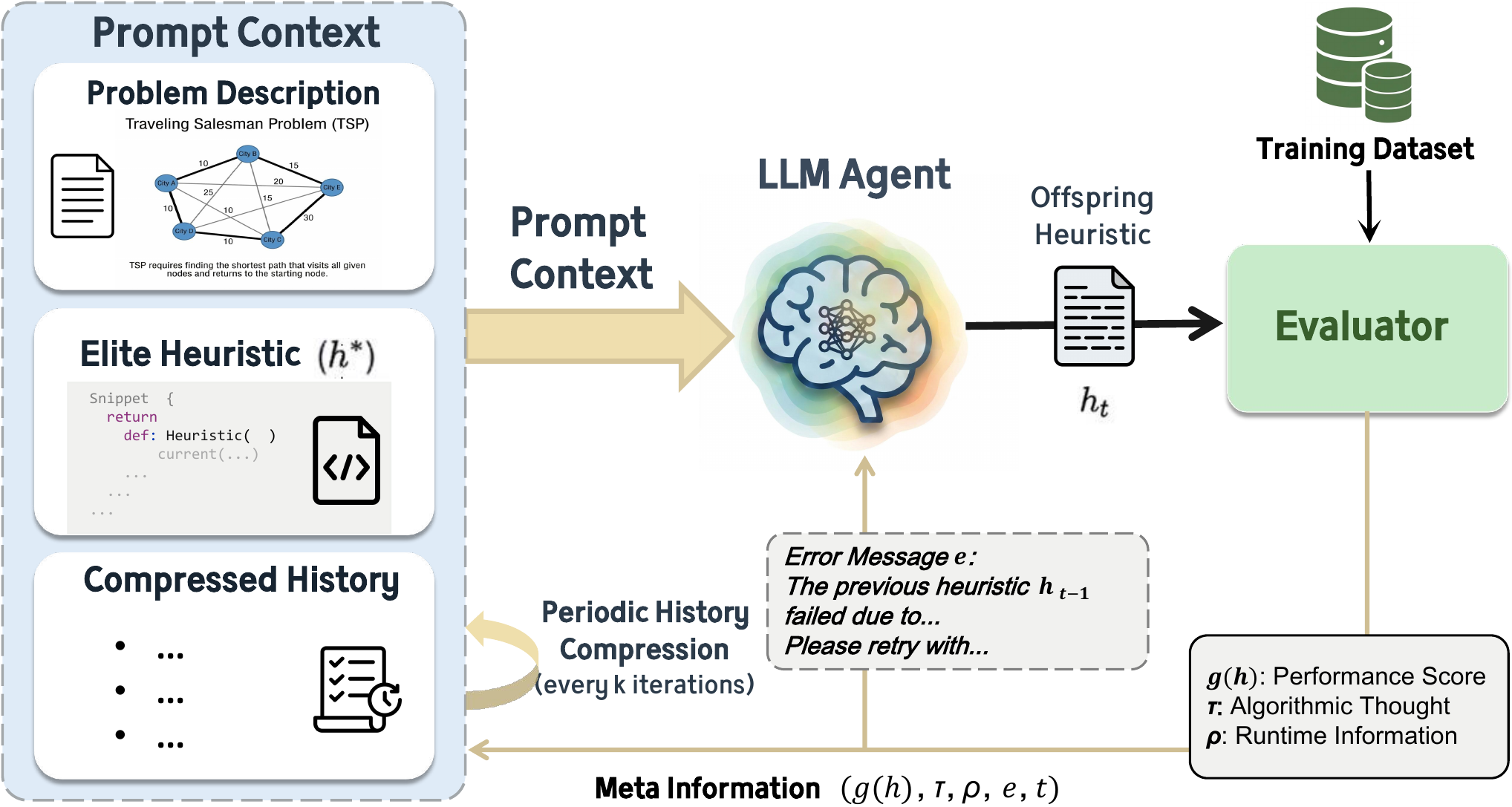}
    \caption{Overview of SimpleEvol. 
    }
    \label{fig:simpleevol_framework}
    \vspace{-10pt}
\end{wrapfigure}
\textbf{Framework overview}.
As illustrated in Figure~\ref{fig:simpleevol_framework}, SimpleEvol maintains only a \textbf{single evolving heuristic trajectory}. 
At each iteration, the LLM is called to generate one offspring heuristic together with a concise natural-language description $\mathcal{T}$. 
The offspring heuristic is then evaluated on the target training problems, producing meta information $\xi$ such as objective value, execution time, and possible error messages, which is appended to the running context for the next iteration. 
Formally, let $h_t$ denote the current heuristic at iteration $t$. 
SimpleEvol alternates between two steps:


\begin{itemize}
    \item \textbf{Generation:} the LLM proposes a new heuristic $h_{t}$ conditioned on the task description and a compact history of previous attempts, until a maximum budget $T$ is reached;
    \item \textbf{Evaluation:} $h_{t}$ is executed on the training dataset to obtain meta information $\xi_t = (g(h_t), \mathcal{T}_t, \rho_t, e_t, t)$, which is then converted to textual context for the next iteration step. 
\end{itemize}

\textbf{History compression and memory}.
A key challenge in feedback-driven LLM-based search is that the context grows rapidly with the number of generations. 
To control this cost, SimpleEvol periodically performs history compression. Every $k$ iterations, the previous meta information, together with the current best-so-far candidate, is summarized into concise working notes that encompass the key experimental histories, including successful or ineffective structural patterns and common implementation errors. 
This compressed memory is then retained in the context while older records are removed. Although intentionally simple, the memory mechanism allows the framework to accumulate useful search experience across iterations without introducing a separate planner or reflection module.

\textbf{Why this design}?
Guided by the bitter lesson, a high-ICE AHD framework should contain minimal human-designed structure and instead rely on the LLM's autonomous capacity to propose, implement, verify, and refine optimization ideas. The design of SimpleEvol directly follows this philosophy, providing only the bare scaffolding necessary for stable iterative improvement. 
This design has three properties: (1) \textit{low framework complexity}, as it removes explicit population management and multi-operator coordination; (2) \textit{full intelligence utilization}, as model intelligence is leveraged through open-ended planning, trajectory-based reflection, and self-directed generation rather than constrained by predefined rule-based modifications; and (3) \textit{traceability}, as the entire search process is recorded as a sequence of heuristic states, giving the LLM access to recent experiments together with compressed summaries of earlier trajectories. This trace allows the model to identify recurring failure patterns and promising directions in future experiments, supporting more informed self-directed exploration.

These properties make SimpleEvol a natural testbed for our bitter-lesson-style view of AHD. If stronger LLMs are increasingly capable of reasoning, coding, and self-improvement, then a simpler framework with fewer structural constraints and access to the search trajectory should convert these capabilities into optimization performance more effectively than a heavily engineered pipeline. Consequently, it would exhibit better scalability with model intelligence.
In the experiments, we evaluate whether, despite its simplicity, SimpleEvol can remain competitive and provide empirical evidence for the proposed AHI--ICE perspective. Detailed prompts used in SimpleEvol are collected in Appendix \ref{sec: Prompts_used}.



\section{Experiments}
\label{sec:experiments}

 \textbf{Evaluated Optimization Tasks.} We conduct our analysis on three well-studied optimization tasks that are widely adopted in existing LLM-based AHD research ~\cite{romera2024mathematical,liu2024evolution,ye2024reevo}:

\textit{\underline{1) Traveling Salesman Problem (TSP).}} TSP is formulated as finding a minimum-length tour that visits each node exactly once and returns to the starting node \cite{matai2010traveling}. 
We consider the step-by-step constructive framework, where the evolved heuristic is used to select the next node conditioned on the current partial tour \cite{ye2024reevo}. 
\textit{\underline{2) Capacitated Vehicle Routing Problem (CVRP).}} The objective of CVRP~\cite{toth2014vehicle} is to design a set of routes to serve all customers under vehicle capacity constraints with minimum total travel length. 
We consider the ant colony optimization (ACO) framework \cite{dorigo2006ant}, where the evolved heuristic is used to generate the heuristic information matrix, which is combined with pheromone values to guide the edge selection during solution construction.
\textit{\underline{3) Flow Shop Scheduling Problem (FSSP).}} The objective is to find a job sequence of length $n$ that minimizes the makespan on $m$ machines~\cite{emmons2012flow}. There are $m$ operations for each job to be performed in a predefined order on these machines. Under the GLS framework, the evolved heuristic is used to update the search landscape and select jobs for perturbation during local search.
We follow the same procedures as previous studies~\cite{liu2024evolution,zheng2025monte} to generate the datasets of these problems. They cover three representative metaheuristic settings, i.e., constructive search, ACO, and GLS, across two major CO domains (routing and scheduling), providing a sufficiently diverse benchmark for our cross-framework analysis. 

\textbf{Implementation Details.} Our study spans 10 LLMs covering both reasoning and non-reasoning models across all three evaluated tasks. The selected models include \OpenAIIcon models (\texttt{GPT-4o-mini}~\cite{hurst2024gpt}, \texttt{GPT-4.1-nano}, \texttt{GPT-4.1-mini}, \texttt{o3-mini}, and \texttt{GPT-5-mini}~\cite{singh2025openai}), 
\AnthropicIcon model (\texttt{Claude Sonnet 3.7}~\cite{anthropic2025claude37}), 
\GoogleIcon model (\texttt{Gemini-2.5-Flash}~\cite{comanici2025gemini}), 
\DeepSeekIcon model (\texttt{DeepSeek-v3}~\cite{liu2024deepseek}), 
and \QwenIcon models (\texttt{Qwen3-235B-Instruct}~\cite{yang2025qwen3} and \texttt{Qwen3-235B-Thinking}). 
Among them, \texttt{o3-mini}, \texttt{GPT-5-mini}, and \texttt{Qwen3-235B-Thinking} belong to the reasoning-model family.
For each (AHD method, model) pair, we conduct 3 independent runs and average the objective scores on each problem size. For TSP, the study corresponds to approximately 118k LLM calls in total, with an aggregated API cost of about $\$1272$. Across all experiments, we fix the maximum number of generated heuristics to 820 to ensure fairness following EoH \cite{liu2024evolution}. Full descriptions of experimental implementations are in Appendix \ref{sec:experimental_setups}.

\textbf{Compared Baselines.} We mainly compare four LLM-based AHD frameworks with different levels of framework complexity, namely FunSearch \cite{romera2024mathematical}, EoH \cite{liu2024evolution}, ReEvo \cite{ye2024reevo} and our SimpleEvol. To further broaden the framework coverage of the AHI--ICE analysis, we additionally
evaluate MCTS-AHD~\cite{zheng2025monte} as a structurally distinct tree-search framework, and the corresponding AHI and ICE results are reported in Appendix~\ref{sec:ICE_variants}. Among them, FunSearch is the earliest representative framework in this line of work, while ReEvo is a more recent and more structurally sophisticated framework. For TSP, we report test performance by the optimality gap to reference solutions computed by LKH-3, a widely used near-optimal TSP solver \cite{helsgaun2017extension}. For CVRP, we follow MCTS-AHD \cite{zheng2025monte} and use the solutions produced by DeepACO \cite{ye2023deepaco} as reference values. For FSSP, synthetic training instances use the lower bound for the gap computation. 


\subsection{Intelligence conversion across AHD frameworks}
\label{sec:ice_main_results}

\begin{figure}[t]
    \centering
    \includegraphics[width=0.9\linewidth]{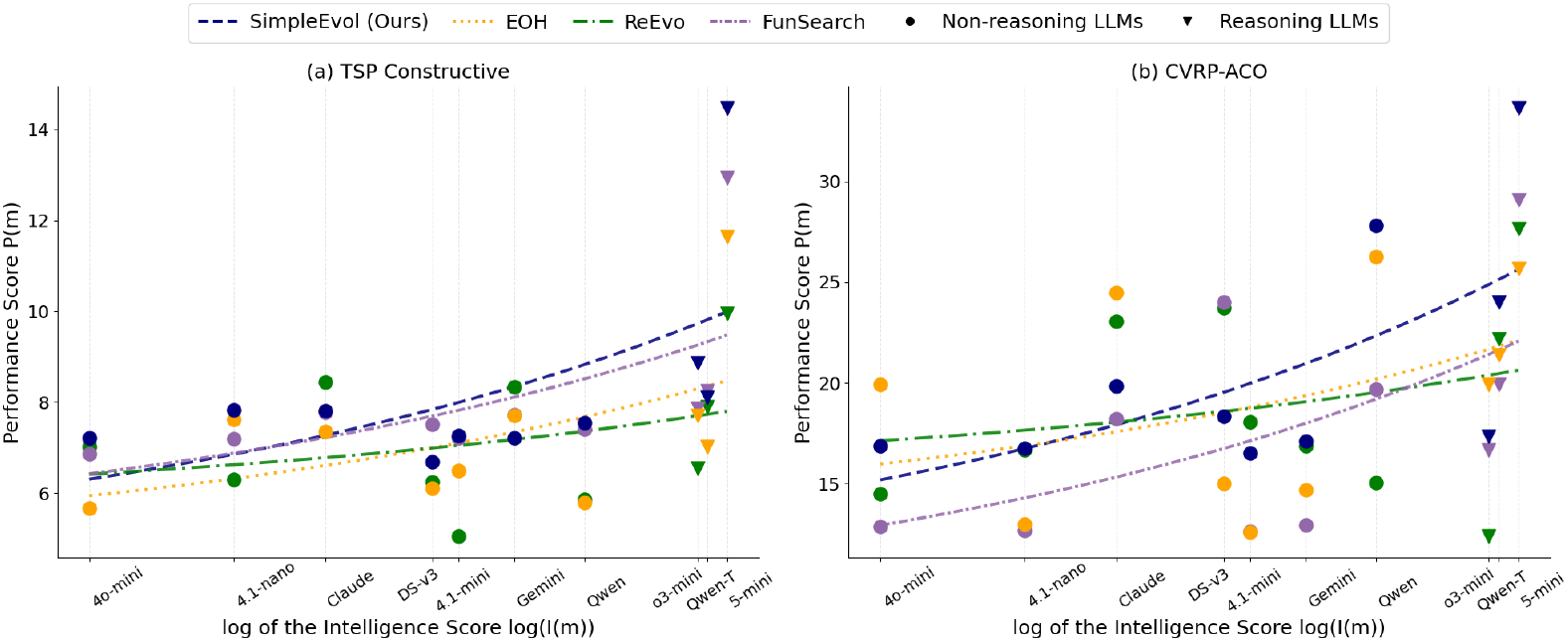}
    \caption{Relationship between model intelligence and AHD performance on TSP and CVRP.}
    \label{fig:pi_plot}
\end{figure}

Figure~\ref{fig:pi_plot} shows how $P(m)$ varies with $I(m)$ on TSP Constructive and CVRP-ACO. Although the relationship is not monotone at the individual model level, the overall trend is consistently positive across both tasks and stronger models tend to produce better heuristics, confirming that LLM intelligence is an important driver of heuristic quality. The fitted trends therefore provide a global summary of how each framework responds to increasing model capability despite local fluctuations across individual backbone models.
SimpleEvol exhibits the steepest and most consistent fitted trend on both tasks, showing that its lightweight structure is more effective at converting model intelligence into optimization performance and more robust to model-level variability. 

\begin{wraptable}{r}{0.48\textwidth}
\vspace{-10pt}
\centering
\small
\caption{ICE on TSP Constructive and CVRP-ACO, together with handcraftedness index (AHI).}
\label{tab:ice_regression_main}
\setlength{\tabcolsep}{3.5pt}
\renewcommand{\arraystretch}{1.08}
\begin{tabular}{lccc}
\thickhline
\thickhline
Method & AHI & ICE(TSP) & ICE(CVRP) \\
\hline
FunSearch         & 6.915 & \underline{1.8174} & \underline{5.4381} \\
EoH               & 8.919 & 1.5082 & 3.6572  \\
ReEvo             & 9.112 & 0.8230 & 2.0824 \\
SimpleEvol (ours) & 6.001 & \textbf{2.1941} & \textbf{6.2128} \\
\thickhline
\thickhline
\end{tabular}
\vspace{-10pt}
\end{wraptable}
Table~\ref{tab:ice_regression_main} quantifies this trend using regression-based ICE, reported alongside AHI for each framework. On both tasks, SimpleEvol achieves the highest ICE despite having the lowest AHI, while ReEvo obtains the lowest ICE despite having the highest AHI.  This contrast is consistent across the two problem settings and reveals a clear separation between framework complexity and intelligence conversion efficiency. 
The inverse relationship between ICE and AHI supports our central hypothesis that increasing framework complexity does not necessarily improve intelligence conversion efficiency. Instead, lighter AHD frameworks appear better positioned to preserve and exploit the performance gains enabled by stronger backbone models, allowing increases in model intelligence to translate more directly into improved heuristic quality.

\begin{figure*}[b]
\centering

\begin{subfigure}[h]{0.495\textwidth}
    \centering
    \includegraphics[width=\linewidth]{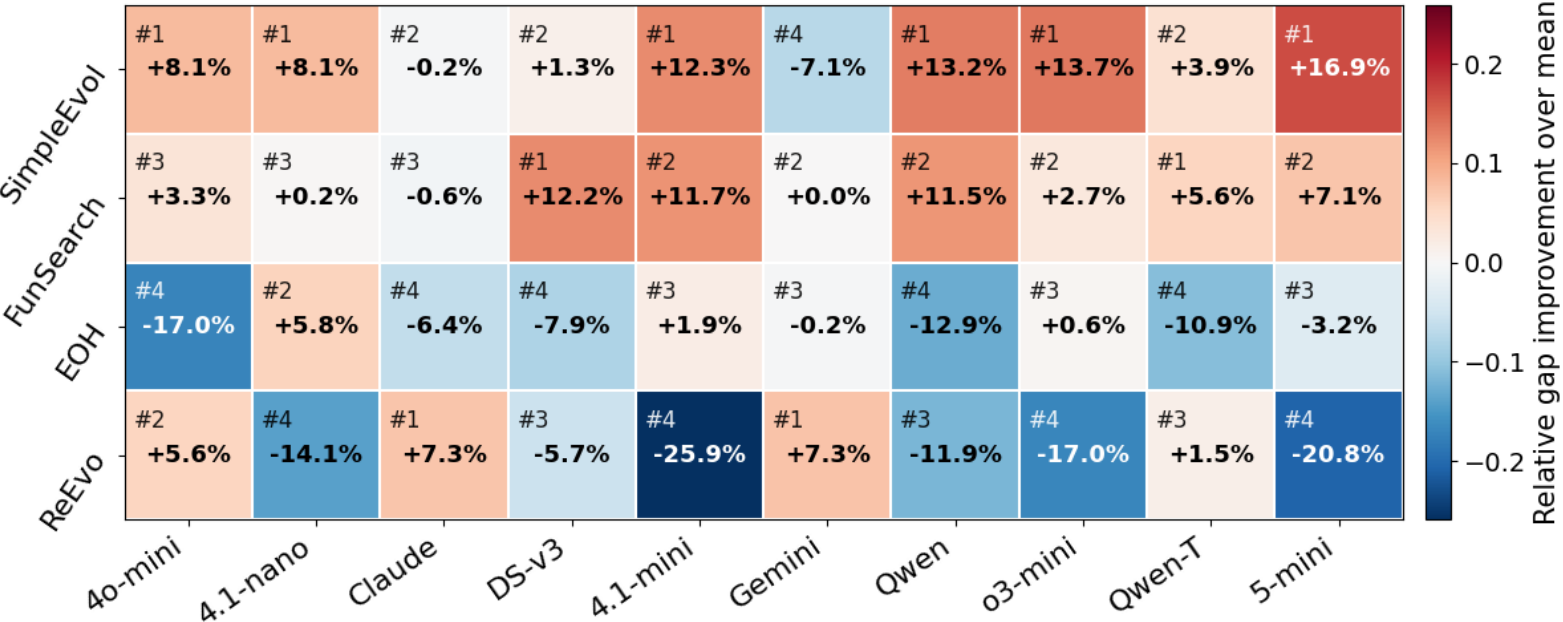}
    \caption{TSP Constructive}
    \label{fig:tsp_adv_rank}
\end{subfigure}
\hfill
\begin{subfigure}[h]{0.495\textwidth}
    \centering
    \includegraphics[width=\linewidth]{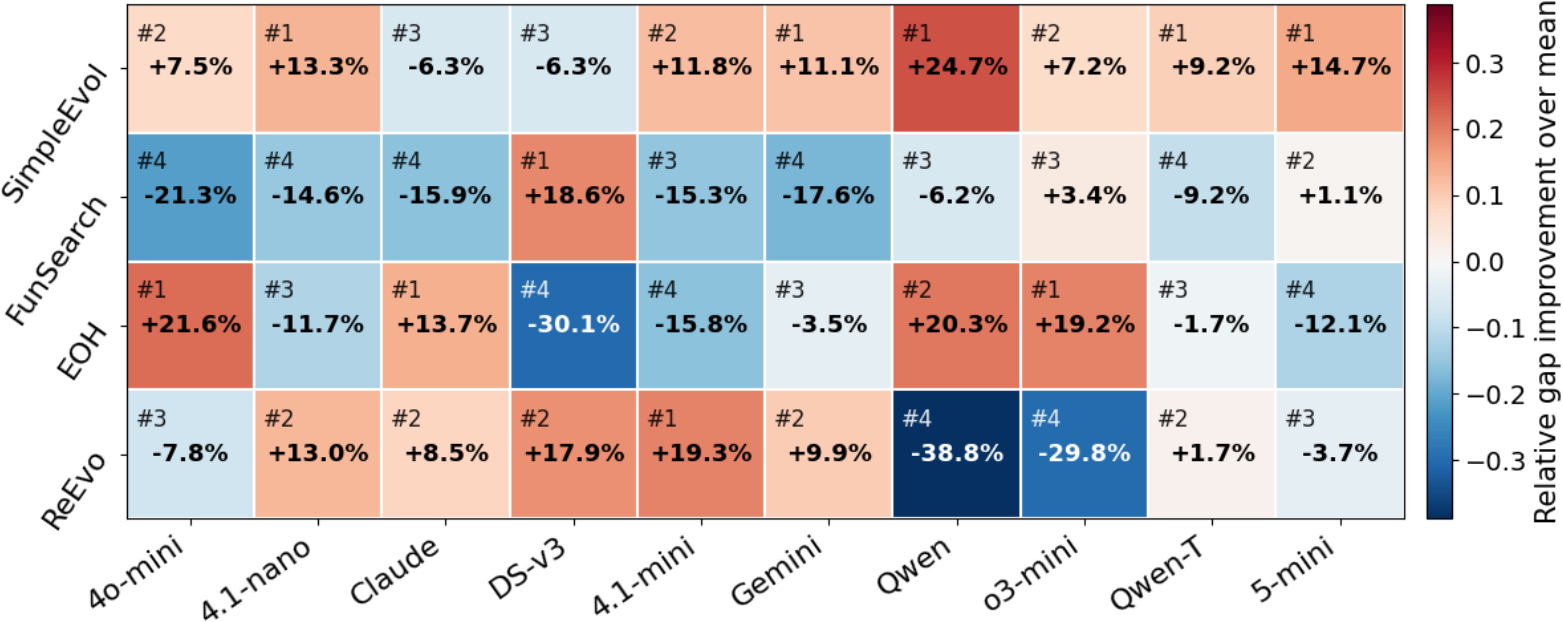}
    \caption{CVRP-ACO}
    \label{fig:cvrp_adv_rank}
\end{subfigure}

\caption{
Relative framework advantage under different model intelligence. The upper-left number in each cell denotes the performance rank on each backbone model.
}
\label{fig:advantage_rank_dual}
\end{figure*}

\textbf{Relative dominance under fixed model intelligence}. Beyond the global ICE trend, Figure~\ref{fig:advantage_rank_dual} compares AHD frameworks under each fixed backbone model. 
For every model, frameworks are ranked by average gap, while cell color and centered values indicate relative gap improvement over the mean across all methods under the same model. 
This within-model view removes cross-model scale effects and directly measures which framework better exploits a given model intelligence.
Notably, while remaining competitive on the weaker models, SimpleEvol increasingly occupies the top rank as model intelligence grows, particularly on CVRP-ACO. 
It also shows stronger positive relative gap improvement in the high-intelligence regime, indicating that its higher ICE is not driven only by the fitted regression trend, but is also reflected in direct model-wise dominance. Table~\ref{tab:gpt5mini_cross_framework} further demonstrates the superior absolute performance of SimpleEvol under the strongest reasoning backbone, \textbf{GPT-5-mini}, where it consistently achieves the lowest gaps across all settings.
\begin{table*}[t]
\captionsetup{labelfont=bf}
\small
\centering
\caption{Performance comparison of AHD frameworks using \textbf{GPT-5-mini}. Step-by-step construction is abbreviated as SC. FunSearch~\cite{romera2024mathematical} is abbreviated as Fun. The best result of each setting is in bold.}
\label{tab:gpt5mini_cross_framework}
\setlength{\tabcolsep}{1.5pt}
\renewcommand{\arraystretch}{1.12}
\begin{tabular}{l|cccc|cccc|cccc}
\noalign{\hrule height 1.2pt}
\multirow{2}{*}{Task}
& \multicolumn{4}{c|}{N=50}
& \multicolumn{4}{c|}{N=100}
& \multicolumn{4}{c}{N=200} \\
\cline{2-13}
& Fun. & EoH & ReEvo & Ours
& Fun. & EoH & ReEvo & Ours
& Fun. & EoH & ReEvo & Ours \\
\hline
TSP-SC.
& 5.50\% & 6.76\% & 7.98\% & \textbf{4.77\%}
& 7.33\% & 8.44\% & 10.06\% & \textbf{6.47\%}
& 10.35\% & 10.56\% & 12.10\% & \textbf{9.49\%} \\
CVRP-ACO
& 1.04\% & 0.71\% & 0.49\% & \textbf{0.34\%}
& 4.83\% & 6.70\% & 5.98\% & \textbf{4.31\%}
& 4.46\% & 4.28\% & 4.38\% & \textbf{4.26\%} \\
\noalign{\hrule height 1.2pt}
\end{tabular}
\end{table*}

\subsection{Extension to FSSP-GLS and other benchmarks}
\label{sec:more_benchmarks_fssp}
We further evaluate SimpleEvol on FSSP under the GLS framework and on TSPLIB instances in Appendix~\ref{sec:extension_to_fssp_gls} and \ref{sec:result_on_tsplib}. On FSSP, SimpleEvol achieves the highest ICE on the in-distribution set and the best performance under 8 of the 10 backbone models on the Taillard benchmark. On TSPLIB, it also achieves the best average gap and the largest number of top-1 results. These results further demonstrate the effectiveness and generalization ability of SimpleEvol across additional problem settings and benchmarks.


\subsection{Cost Efficiency Analysis}
\label{sec:main_cost_analysis}

\begin{wrapfigure}{r}{0.50\textwidth}
\vspace{-8pt}
\centering

\begin{minipage}{\linewidth}
\captionsetup{type=table}
\centering
\caption{Ablation study of SimpleEvol on TSP and CVRP at size 50, averaged across 5 independent runs. Each entry reports the gap to the reference objective, with objective-value standard deviation in parentheses. Lower gap is better.}
\label{tab:ablation_main}
\small
\setlength{\tabcolsep}{2.2pt}
\renewcommand{\arraystretch}{1.08}
\begin{tabular}{l|cc}
\thickhline
\thickhline
Method & \textit{TSP50}  & \textit{CVRP50} \\
\hline
SimpleEvol (Default) 
& \textbf{10.00\%} {\scriptsize(0.1020)}
& \textbf{3.08\%} {\scriptsize(0.1643)} \\
\hline
w/o summary 
& 13.24\% {\scriptsize(0.0711)}
& 7.38\% {\scriptsize(0.2826)} \\

w/o meta information 
& 11.67\% {\scriptsize(0.0477)}
& 6.01\% {\scriptsize(0.1243)} \\

w/o best-of-so-far 
& 14.56\% {\scriptsize(0.0036)}
& 11.49\% {\scriptsize(0.1345)} \\

compress every 10 
& 12.88\% {\scriptsize(0.0942)}
& 6.01\% {\scriptsize(0.2381)} \\

\thickhline
\thickhline
\end{tabular}
\end{minipage}

\vspace{4pt}

\begin{minipage}{\linewidth}
\centering
\includegraphics[width=0.98\linewidth]{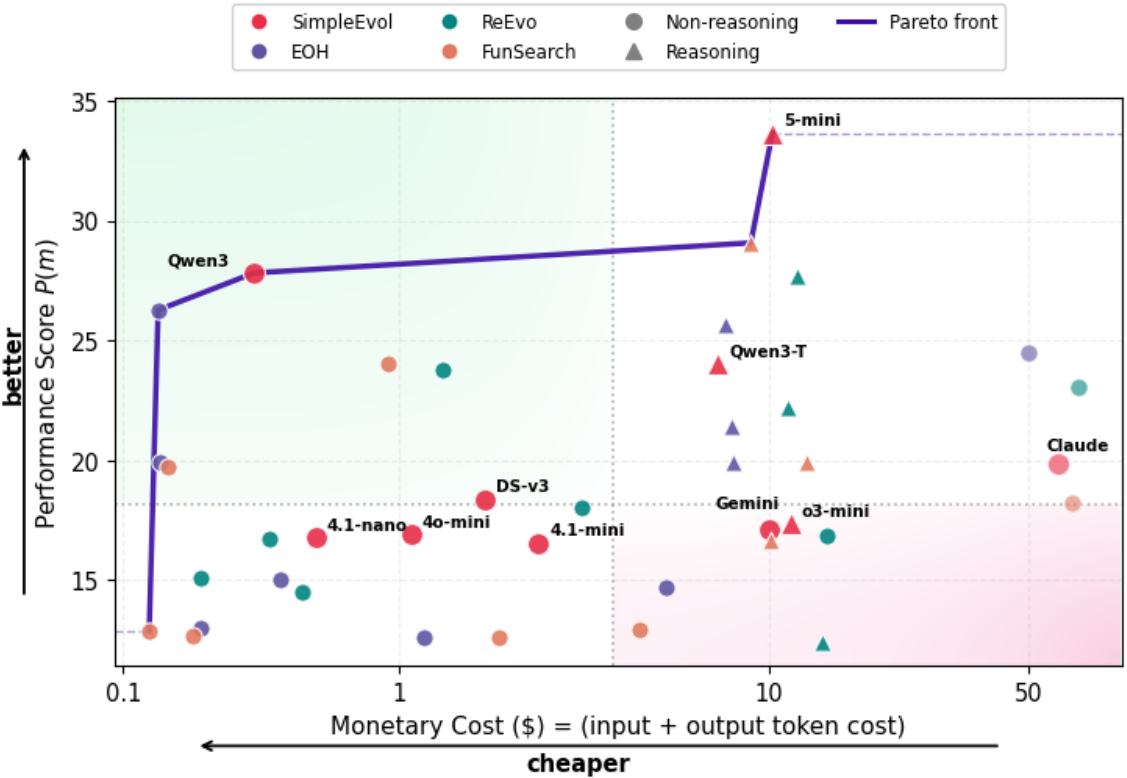}
\caption{Cost-performance balance on CVRP-ACO. The Pareto frontier is highlighted in blue.
}
\label{fig:cost_pareto}
\end{minipage}

\vspace{-10pt}
\end{wrapfigure}
Figure~\ref{fig:cost_pareto} plots the performance score $P(m)$ against the total API cost for all (framework, model) pairs on CVRP-ACO, where the cost is the sum of input and output token charges under each model's pricing. The two dotted lines denote the median value of each dimension. 
It shows that SimpleEvol occupies a favorable position in this cost–performance space. Two of its model instances, namely Qwen3-235B-Instruct and GPT-5-mini, lie on the Pareto-optimal frontier, indicating that no other framework achieves both lower cost and higher performance simultaneously under these models. Several additional SimpleEvol instances, including DeepSeek-v3 and Qwen3-235B-Thinking, fall near the frontier, clustering in the upper-left region of the plot.

We also include a detailed breakdown of runtime, input and output token usage, and LLM query counts across all models in Appendix~\ref{sec:cost_analysis}. The results show that despite SimpleEvol's use of periodic history summarization and richer meta-information as context, its overall monetary cost remains well-controlled and comparable to other AHD frameworks. This suggests that the additional input token consumption introduced by the feedback-driven design does not translate into disproportionate expenditure. 


\subsection{Ablation Study}
\label{sec:main_ablation}







We conduct an ablation study to examine the contribution of key components in SimpleEvol using GPT-4.1-nano, including summarization, meta-information, best-of-so-far retention, and compression frequency. 
As shown in Table \ref{tab:ablation_main}, removing the best-of-so-far heuristic leads to the largest performance drop, indicating the importance of maintaining an elite heuristic as a strong structural prior during the search. 
Similarly, removing summarization or meta-information also results in noticeable degradation, suggesting that compressed trajectory information plays a key role in guiding subsequent heuristic generation. Increasing the compression interval also leads to performance degradation, showing that less frequent summarization forces the model to process longer histories, introducing more noise and diluting attention away from the most informative structural patterns. Overall, these results confirm that the design of SimpleEvol is not the result of arbitrary simplification but rather a combination of lightweight yet essential components that together enable effective intelligence conversion.

\section{Conclusion}
\label{sec:conclusion}



This work systematically examines how framework-level complexity influences the conversion of model intelligence into optimization performance in LLM-based AHD. We introduce AHI to quantify human priors and ICE to measure how effectively an LLM-driven framework translates gains in model intelligence into heuristic quality. Across ten backbone LLMs and three combinatorial optimization problems, we find a consistent inverse relationship between AHI and ICE. SimpleEvol, our minimal agent-loop framework with almost no human-designed priors, consistently achieves the highest ICE on all primary benchmarks while remaining cost-competitive. These results challenge the prevailing trend of increasingly complex AHD pipelines and point to a lighter, more model-centric alternative with less human priors. A natural implication is that future frameworks should grant LLMs greater autonomy over the full search process, including planning, reflection, and memory, rather than embedding them inside elaborate outer loops. As foundation models continue to advance, the frameworks best positioned to exploit these gains will be those that get out of the model's way.

\begin{ack}
This research is supported by the National Research Foundation, Singapore under its AI Singapore Programme (AISG Award No: AISG3-RP-2025-036-USNSF). We thank the anonymous reviewers and the area chair for valuable discussions and feedback. 
\end{ack}

{
\small

\input{references}
}
\newpage

\appendix
\clearpage
\startcontents[appendix]

\section*{Appendix Contents}
\setcounter{tocdepth}{2}
\printcontents[appendix]{}{1}{}

\newpage
\section{Related Work}
\label{sec:related_work}

\paragraph{Automated Heuristic Design.}
Automated Heuristic Design (AHD), also closely connected to the broader literature on hyper-heuristics and automated algorithm design, studies how to automatically construct, adapt, combine, or select heuristics for challenging search and optimization problems \cite{burke2013hyper, burke2018classification, pillay2018hyperheuristics, qu2020general}. Earlier work in this area explored genetic programming, grammatical evolution, component-wise algorithm design, and automatic configuration frameworks as ways to search over spaces of heuristic procedures rather than hand-crafting them case by case \cite{langdon2002foundations, oneill2002grammatical, hutter2009paramils, bezerra2015automatic, stutzle2018automated}. A common theme across these studies is that algorithm design itself can be cast as a higher-level optimization problem. At the same time, these methods also reveal a longstanding tension that stronger search performance is often obtained by introducing richer representations, more search operators, or more carefully engineered control logic, which can increase the amount of human-designed structure embedded in the framework \cite{branke2015automated,qu2020general, camacho2023designing}. Our work is related to this line in that it also treats heuristic design as a system-level problem. However, rather than proposing another search mechanism, we revisit AHD from a different angle by characterizing the complexity of AHD frameworks themselves and study what that complexity implies for how effectively a framework exploits improvements in the underlying language model.

\paragraph{LLM-based Automated Heuristic Design.}
Recent large language models (LLMs) have substantially expanded the scope of AHD by enabling heuristic generation directly in code space, often with natural-language reasoning, iterative refinement, and black-box evaluation \cite{romera2024mathematical, liu2024evolution, ye2024reevo, dat2025hsevo, zheng2025monte, liu2026systematic}. This line of work has shown that LLMs can serve not only as code generators, but also as search operators, reflectors, and controllers inside automated heuristic evolution pipelines. Importantly, progress in this literature has largely been driven by \emph{framework engineering}: researchers design increasingly sophisticated outer loops involving population management, crossover and mutation variants, reflection modules, memory, diversity control, tree search, or optimizer-level meta-search \cite{liu2024evolution, ye2024reevo, dat2025hsevo, zheng2025monte, shi2026generalizable, chen2025hifo, liu2026eoh, xie2025llm}. These advances have produced stronger empirical performance, but they also make methods harder to compare purely on the basis of model intelligence, since performance differences increasingly reflect both the LLM and the manually designed orchestration wrapped around it. In this sense, existing LLM-based AHD methods are not only heuristic search methods; they are also different hypotheses about how much external structure should be imposed on the model. Our work is motivated by the fact that this second aspect remains under-studied. We therefore shift the focus from designing a more elaborate heuristic evolution pipeline to analyzing the complexity of such pipelines and asking whether additional framework structure necessarily leads to better utilization of stronger models. A related trend is seen in open-ended discovery, where CORAL~\cite{qu2026coral} delegates search decisions to autonomous agents rather than relying on fixed evolutionary scaffolds.  While CORAL studies autonomous multi-agent program evolution, our work focuses on LLM-based AHD and quantitatively analyzes how framework complexity affects the conversion of model intelligence.

\paragraph{Neural Combinatorial Optimization as a Contrasting Paradigm.}
A related but conceptually distinct direction is Neural Combinatorial Optimization (NCO), where neural networks are trained to construct or improve solutions directly from data \cite{bengio2021machine, bello2016neural, nazari2018reinforcement, kool2018attention, kwon2020pomo, luo2023neural, drakulic2023bq, drakulic2024goal}. Compared with LLM-based AHD, NCO typically places more of the problem-solving burden inside the learned model itself, rather than in a symbolic outer-loop that repeatedly rewrites and evaluates heuristic code at test time. This makes NCO an informative reference point for our study. It highlights a broader design question in neural combinatorial optimization: when performance improves, is the gain primarily driven by increasingly elaborate architectural choices, carefully designed training pipelines, or reward and objective designs? Prior work has investigated generalization, scalability, data-driven learning, and
synthetic-to-real adaptation in NCO \cite{bengio2021machine, manchanda2022generalization, gao2023towards, berto2024routefinder, zhou2024mvmoe,zhu2026bridging}, but comparatively less attention has been paid to a parallel question of which LLM-AHD framework designs best preserve and translate the improvements of foundation models into downstream optimization performance. Our work addresses this question by positioning LLM-based AHD frameworks along a complexity axis, thereby complementing existing comparisons based solely on final objective value.

\paragraph{The Bitter Lesson, scaling, and intelligence conversion.}
Our perspective is also inspired by a broader lesson from AI research that systems that achieve the strongest long-term progress often rely less on handcrafted domain-specific structure and more on scalable learning and computation \cite{sutton2019bitter}. In the LLM era, this observation has become especially salient, as model intelligence has improved dramatically with scale, data, and training efficiency, leading to broad gains in language understanding, coding, and reasoning \cite{brown2020language, kaplan2020scaling, hoffmann2022training}. Yet in LLM-based AHD, stronger models are usually treated as drop-in replacements inside existing frameworks, leaving the question of \emph{intelligence conversion} largely implicit. It remains unclear how much additional model intelligence is actually converted into better heuristic search outcomes when a stronger model is substituted into an AHD system. This question is central to our paper. Rather than evaluating AHD frameworks only by their final best performance on a single LLM backbone, we argue that they should also be examined in terms of how much structural complexity they introduce and how efficiently they transform model-side intelligence improvements into downstream optimization gains. From this perspective, our work is not another proposal for a more complicated search scaffold. Instead, it offers a complexity-aware view of progress in LLM-based AHD, motivated by the possibility that, beyond some point, increasing framework complexity may weaken rather than strengthen the effective use of model intelligence.

\section{Prompts Used in SimpleEvol}
\label{sec: Prompts_used}

\textbf{System Prompt.} The system generator prompt delineates the core experimental protocols for LLM-based automatic heuristic search. It instructs the model to iteratively design and evaluate candidate heuristics. More importantly, rather than prescribing specific operations on given parent heuristics (e.g., explicitly combining two candidates) in each iteration, the prompt casts LLMs as a planner that designs the whole experimental process for discovering more effective heuristics. 

Specifically, the model is encouraged to maintain an internal loop consisting of documenting past attempts, reflecting on their outcomes (e.g., errors), comparing with previous strategies, and planning subsequent improvements. By explicitly organizing the reasoning process into “record–reflect–compare–plan” stages, the prompt encourages LLMs to accumulate knowledge throughout experiments and to adapt its search strategy based on meta information and summary. As shown in Fig. \ref{prompt:system generator prompt}, the prompt is generated by inserting only the number of maximum experiments. Apart from defining the general requirements, the prompt also describes the roles of the generator LLM as a summary writer or a code designer.

\tcolorboxenvironment{promptbox}{
  colback=gray!10,     
  colframe=gray!60,    
  boxrule=0.5pt,
  arc=3pt,             
  left=6pt,
  right=6pt,
  top=6pt,
  bottom=6pt
}

\begin{figure}[t]
\centering
\begin{tcolorbox}[colback=green!3!white,colframe=black,boxrule=0.8pt,arc=4pt,width=\linewidth]
{\sffamily
\small
\setstretch{1.1}
\raggedright
\textls[20]{%
You are an algorithm optimization expert specializing in heuristic design for combinatorial optimization problems. Your task is to explore and optimize constructive heuristics for the given combinatorial optimization problem. Through multiple experiments, you will try to discover better heuristic strategies.

\vspace{0.5em}
\#\# Experiment Requirements

You may perform at most \textcolor{vividPurple}{\{max\_experiments\} }experiments.

After each experiment, you should internally:

\hspace{1.5em} ** \textbf{Record} **: record the heuristic design idea and the result

\hspace{1.5em} ** \textbf{Reflect} **: analyze why the design worked or failed

\hspace{1.5em} ** \textbf{Compare} **: compare with previous experiments

\hspace{1.5em} ** \textbf{Plan} **: think about what improvement to try next

\vspace{0.5em}
    Continue designing new heuristics and testing them through experiments.

    You may form hypotheses and verify them through experiments to gain domain knowledge.
    Not every experiment must improve the result, but experiments should not be meaningless.
    If a certain design direction fails repeatedly, you should abandon it and explore different ideas.

\vspace{0.5em}
    \colorbox{rosePink!15}{1. When you are doing summarization:}
    All summaries are concise working notes for future exploration (previous context will be cleared after summarization to save tokens).
    Therefore, avoid unnecessary wording and focus only on conclusions, insights, and hypotheses that influence future exploration.
    
        
        
        
        

\vspace{0.5em}
    \colorbox{rosePink!15}{2. When you are designing code:}
    
    \#\#\# Coding Rules 
    
    - Output Python code and description.
    
    - The required function signature will be provided in the problem description.
    
    - Write all necessary imports inside functions (not at the top level of the file). 
    
    - Ensure correct Python indentation. Do not add extra indentation outside functions. Do not place return statements outside the function.
    
}
}
\end{tcolorbox}
\caption{Template of the system generator prompt.}
\label{prompt:system generator prompt}
\end{figure}

\textbf{Summary Prompt.} The summary user prompt is created using a template as shown in Fig. \ref{prompt: summary_user_prompt}. By instructing LLMs to briefly summarize the previous trajectory of all experiments, extract the performance profile of attempted design strategies while capturing the recurring failure patterns, the prompt attempts to transform the previous history information into reusable accumulation of experience. Moreover, the summary is designed to be a descriptive and non-prescriptive abstraction of the search trajectory, instilling learnable expertise and the awareness of the evolution process into LLMs. 

\textbf{Summary Injection Prompt.} The summary injection prompt shown in Fig. \ref{prompt:summary_assistant_prompt} is formatted by inserting the number of completed experiments, the maximum number of experiments, and the summarized context generated during summarization. The summary is then added to the context as an assistant response, serving as a compressed memory of the past trajectory for subsequent reasoning.

\begin{figure}[t]
\centering
\begin{tcolorbox}[
    colback=green!3!white,
    colframe=black,
    boxrule=0.8pt,
    arc=4pt,
    width=\linewidth
]
{\sffamily
\small
\setstretch{1.12}
\raggedright
\textls[20]{%

Please provide an \textbf{**intermediate summary**} of the \textcolor{vividPurple}{\{experiment\_count\}} experiments you have conducted so far, including:

\vspace{0.5em}

\hspace{1.5em} 1. \textbf{**Experiment History**}: List the strategy and result (obj value) of each experiment. Do not discard records from previous summaries. 
    If many experiments use similar strategies and produce similar results, you may group them together.

\vspace{0.5em}

\hspace{1.5em} 2. \textbf{**Key findings**}: How do different strategies perform? Analyze the possible reasons.

\vspace{0.5em}

\hspace{1.5em} 3. \textbf{**Mistakes to avoid**}: Which improvement methods have been proven ineffective, and which coding patterns tend to cause errors.

\vspace{0.5em}

\hspace{1.5em} 4. Based on the previous summaries and the results of these experiments, \textit{what knowledge or insights about this task can we learn?}

\vspace{0.5em}

Please keep the summary concise but complete. The summary should also include the content from the previous summary, since it will serve as the basis for future exploration. Do not provide suggestions for the next step. Do not include full codes in your summary. Keep your summary within 300 words.

}
}
\end{tcolorbox}
\caption{Template of the summary user prompt.}
\label{prompt: summary_user_prompt}
\end{figure}

\begin{figure}[t]
\centering
\begin{tcolorbox}[colback=green!3!white,colframe=black,boxrule=0.8pt,arc=4pt,width=\linewidth]
{\sffamily
\small
\setstretch{1.12}
\raggedright
\textls[20]{%

[Intermediate Summary - \textcolor{vividPurple}{\{experiment\_count\}}/\textcolor{vividPurple}{\{max\_experiments\}} experiments completed]

\vspace{0.6em}

\textcolor{myPink}{\{summary\_content\}}

}
}
\end{tcolorbox}
\caption{Template of the summary injection prompt.}
\label{prompt:summary_assistant_prompt}
\end{figure}

\textbf{User Generator Prompt with Compressed History.} 
At each context compression and summarization step, the user generator prompt in Fig. \ref{prompt:user_generator_prompt_with_history} is reconstructed using the current experimental progress and the meta-information of the elite heuristic. The prompt for meta-information feedback used in this process is illustrated in Fig.~\ref{prompt:meta_feedback}. A short behavior clue (e.g., "experiment with hybrid combinations of different heuristic strategies") for encouraging the behavior diversity of agents is also generated and inserted into the user generator prompt.
The user generator prompt is created and updated only when an elite heuristic is available and the first summary has been generated. 


\begin{figure}[t]
\centering
\begin{tcolorbox}[colback=green!3!white,colframe=black,boxrule=0.8pt,arc=4pt,width=\linewidth]
{\sffamily
\small
\setstretch{1.12}
\raggedright
\textls[20]{%

Based on your intermediate summary, please continue exploring.

Experiment progress: \textcolor{vividPurple}{\{experiment\_count\}}/\textcolor{vividPurple}{\{max\_experiments\}} completed, \textcolor{vividPurple}{ \{remaining\_experiments\}} remaining.

\vspace{0.5em}

\colorbox{rosePink!15}{Reference Heuristic}

Current elite code core logic (Experiment \textcolor{vividPurple}{\{elite\_ID\} }, obj=\textcolor{vividPurple}{\{obj\_score\}}):

\textcolor{vividPurple}{\{filtered\_code\} }

Elite code description: \textcolor{vividPurple}{\{code\_description\}}

 \textcolor{vividPurple}{\{behavior\_clue\}}
\vspace{0.5em}



Output the next candidate using exactly the required format: one python code block followed by a heuristic description.

}
}
\end{tcolorbox}
\caption{Template of the user generator prompt.}
\label{prompt:user_generator_prompt_with_history}
\end{figure}

\begin{figure}[t]
\centering
\begin{tcolorbox}[colback=green!3!white,colframe=black,boxrule=0.8pt,arc=4pt,width=\linewidth]
{\sffamily
\small
\setstretch{1.12}
\raggedright

Your last candidate heuristic has been evaluated.

\colorbox{rosePink!15}{Evaluation result:}
\begin{verbatim}
{
  "experiment": <id>,
  "train_obj": <value>,
  "exec_time": <value>,
  "error": <message or null>
}
\end{verbatim}

Please analyze the result internally and output your next candidate heuristic.

}
\end{tcolorbox}
\caption{Meta-information feedback of heuristic evaluation.}
\label{prompt:meta_feedback}
\end{figure}




\textbf{Initialization Prompt.} The initialization user prompt is used to start generating the first experiment (heuristic) by incorporating the problem name and descriptions into the template, as shown in Fig. \ref{prompt:init_prompt}. The problem name and description injected will then be re-used in the experiments afterward. The detailed descriptions of each evaluated problem and heuristic can be found in Appendix \ref{sec:problem_descriptions} while the task-specific prompts can be found in our supplementary materials.

\begin{figure}[t]
\centering
\begin{tcolorbox}[colback=green!3!white,colframe=black,boxrule=0.8pt,arc=4pt,width=\linewidth]
{\sffamily
\small
\setstretch{1.12}
\raggedright
\textls[20]{%

\textcolor{vividPurple}{\{task\_desc\}}

\vspace{0.5em}

Please start your \textcolor{vividPurple}{\{problem\_name\}} heuristic exploration experiment. 
Begin with a simple baseline method, then try different improvement strategies.

\vspace{0.5em}

Output exactly one candidate heuristic using the following format:}

\begin{verbatim}
```python
# your code here
```
\end{verbatim}

\textls[20]{%
Then provide a concise heuristic description after the code block (you may wrap it using <description> ... </description>). Keep within 50 words.

\vspace{0.5em}
\#\#Description requirements:

-Briefly summarize what structure\/scoring rule is used.

-State the key mechanism and logic.

-Mention important parameters \/ thresholds if any.

-Keep this description informative and structured.
\vspace{0.5em}

Note: Write all necessary imports inside functions (not at the top level of the file).

}
}
\end{tcolorbox}
\caption{Template of the initial user prompt.}
\label{prompt:init_prompt}
\end{figure}

\section{Experimental Setups}
\label{sec:experimental_setups}
 This section provides detailed experimental configurations across reported results, including  working environments, selection of LLMs, dataset construction and the default hyperparameters of the evaluated LLM-based AHD methods.

\subsection{Experimental Implementations}
\label{sec:implementation_details}

Unless otherwise stated, the LLM temperature is set to 1.0 for all models in all experiments. This is consistent with the default LLM temperature used in EoH and FunSearch, although ReEvo requires an increase of 0.3 to promote diversity during initialization. However, a temperature over 1.0 is unstable or not applicable to some models like o3-mini and GPT-5-mini, so we fixed the temperature to 1.0 to ensure consistency of evaluations. Moreover, we fix the maximum number of evaluated heuristics to 820 for all problems across all AHD methods (for EoH, the corresponding number of population is 20 while the population size is 10, counting a total number of evaluations equal to $2\cdot10+20\cdot4\cdot10 = 820$). The default frequency for context compression and summarization is 5. 
Each heuristic is evaluated within 60 seconds on TSP and CVRP \cite{zheng2025monte} and within 60s per instance for FSSP. Experiments are carried out on a workstation powered by an AMD Ryzen 9 5950X CPU. 

\subsection{Selection of Large Language Models}
\label{sec:how_i_select_LLMs}
Both non-reasoning and reasoning models are selected to study how existing AHD methods behave across different levels of benchmarked model intelligence. We select these models based on the following criteria: 1) the selected models can cover a wide range of intelligence; 2) the intelligence levels of models are as different as possible. 3) The benchmark scores of the selected models are available on open leaderboards such as Artificial Analysis~\cite{artificialanalysis2026}.


Specifically, we include seven non-reasoning models, consisting of three models from the GPT series, namely \textit{GPT-4o-mini-2024-07-18} for \texttt{GPT-4o-mini}, \textit{GPT-4.1-nano-2025-04-14} for \texttt{GPT-4.1-nano}, and \textit{GPT-4.1-mini-2025-04-14} for \texttt{GPT-4.1-mini}, as well as four models from other families, namely DeepSeek-v3-0324, Gemini-2.5-Flash (Released Jun 17, 2025), Claude Sonnet 3.7, and Qwen3-235B-A22B-Instruct-2507. We also select three reasoning models, namely o3-mini (-2025-01-31), Qwen3-235B-A22B-Thinking-2507, and GPT-5-mini (\textit{GPT-5-mini-2025-08-07}); among them, GPT-5-mini is trained to reason and produce extended chains of thought before outputting the final answer, which enables it to achieve the highest intelligence score and deliver strong performance on tasks involving multi-step inference and algorithmic problem solving~\cite{srivastava2025beyondbench, singh2025gpt5}.

\subsection{Implementations of Datasets}
\label{sec:baseline_implementations}

\textbf{Step-by-step Constructive for TSP.}
For TSP under the step-by-step constructive framework, the training set consists of 64 randomly generated Euclidean TSP instances with $N=50$ nodes, where node coordinates are sampled uniformly from $[0,1]^2$. The test set contains three independently generated sets of 64 instances each, with problem sizes $N \in \{50,100,200\}$. All AHD methods are evaluated on the same training and test instances under identical random seeds. During heuristic evolution, candidate heuristics are selected based on their performance on the training set, while the final reported results are measured on the three held-out test sets.

\textbf{ACO for CVRP.}
For CVRP under the ACO framework, the training set consists of 10 randomly generated CVRP instances with problem size $N=50$. The test set contains three independently generated sets of 64 instances each, with $N=50$, $100$, and $200$, respectively. During both training and testing, all methods share the same ACO solver configuration, including 30 ants and 100 iterations. In this setting, the evolved heuristic defines the heuristic information matrix used together with pheromone trails to guide ant transitions during route construction. 

\textbf{GLS for FSSP.}
For FSSP under the GLS framework, we follow EoH to generate the training set, which includes 64 randomly generated instances. Concretely, each training instance is randomly generated with a fixed number of 50 jobs, while the number of machines is sampled uniformly from $U[2,20]$. The processing time for each job-machine pair is drawn independently and uniformly at random from $\{1,2,...,100\}$. For testing, we use the same Taillard benchmark instance sets as in EoH. Since ReEvo and FunSearch lack natural realizations of this problem setting, we adapt the evaluation pipeline of FSSP-GLS on these baselines with the same evaluation protocols as SimpleEvol and EoH, including the problem-specific prompts, seed functions, and the datasets.

\subsection{Baseline Implementation}

\textbf{AHD Configurations.} We adhere to the original algorithmic configurations of all baseline LLM-based AHD methods, including hyperparameters such as mutation rate, number of samplings per prompt, number of parents per operator, and the hyperparameters related to cluster sampling. For ReEvo, the population size is initialized at 30 and maintained at 10 in subsequent iterations. For EoH, the population size is set to 10 while the number of populations is fixed to 20. Across all problems, the EoH baseline is obtained from the ReEvo codebase, which provides equivalent implementations of EoH for different problem settings, including TSP constructive and CVRP-ACO. We directly adopt these implementations without modification and evaluate all methods under a unified protocol and solver configuration to ensure fair comparison. This implementation follows the original design of EoH and is consistent with the baseline reported in the ReEvo framework.

\textbf{Reference Values and Other Baselines.} For TSP results, we obtain reference solutions using the Lin–Kernighan heuristic \cite{lin1973effective} implemented in the \texttt{elkai} solver, a widely-used Python implementation of LK-based TSP solvers. For problem sizes up to $N=315$, the solver typically returns provably optimal solutions, while for larger instances it provides high-quality approximate solutions. We report optimality gaps with respect to these reference solutions, following the practice in prior AHD literature \cite{ye2024reevo, zheng2025monte}.

For CVRP-ACO test instances, we adopt the results generated by DeepACO \cite{ye2023deepaco} as reference values. Following DeepACO, ACO solutions are produced using the default configurations in its code, including the consistent number of steps per epoch and training epochs, as well as ant population size and graph sparsification which adjust according to different instance sizes. We report performance gaps with respect to these reference values.

For FSSP experiments on Taillard instances, the optimal or best-known reference values are taken directly from the benchmark dataset introduced in \cite{taillard1993benchmarks}. We use the relative makespan gap with respect to these reference values to calculate the performance score $P(m)$ in the OOD setting.

\subsection{Search Framework Configurations}
\label{sec:search_framework_configs}

\begin{table}[h]
\centering
\caption{Search framework hyperparameters for different problem settings.}
\label{tab:search_framework_configs}
\begin{tabular}{l l c c}
\toprule
Framework & Problem & $D_{\text{train}}$ & $D_{\text{test}}$ \\
\midrule
GLS & FSSP & Number of Iterations: 1000 & Same as $D_{\text{train}}$ \\
ACO & CVRP & Number of Ants: 30; Number of Iterations: 100 & Same as $D_{\text{train}}$ \\
\bottomrule
\end{tabular}
\end{table}

We summarize the key hyperparameters of the search frameworks in Table~\ref{tab:search_framework_configs}, where $D_{\text{train}}$ and $D_{\text{test}}$ denote the synthetic training datasets and separate testing datasets, respectively. The listed parameters correspond to the main search budgets or control variables of each framework. For GLS on FSSP, the maximum number of iterations determines the search depth. For ACO on CVRP, the number of ants and iterations control the exploration scale and convergence behavior. 

\section{Details of the Problems Evaluated and the Seed Heuristics}
\label{sec:problem_descriptions}
\textbf{Travelling Salesman Problem (TSP) under Step-by-Step Construction.} The classical TSP requires constructing a minimum-length tour that visits each node exactly once and returns to the starting node. Specifically, we consider symmetric TSP, where there is a pairwise equal distance between every two cities.
We adopt a constructive setting in which the solution is built incrementally, where the evolved heuristic acts as a decision function that selects the next node at each step.
The function takes as input the current node, the destination node, the set of unvisited nodes, and the distance matrix that encodes pairwise distances. 
It outputs the index of the next node to visit.

\textbf{Capacitated Vehicle Routing Problem (CVRP) under ACO.}
The objective of CVRP is to design a set of routes starting and ending at a depot to serve all customer nodes with minimum total travel cost. 
Each customer is associated with a demand and each vehicle has a fixed capacity constraint, requiring that the total demand served along a route does not exceed the vehicle capacity. 
Solutions are constructed by sequentially assigning nodes to routes while respecting capacity constraints, with routes returning to the depot when necessary.

We adopt an Ant Colony Optimization (ACO) framework, in which multiple agents iteratively construct solutions by sampling edges according to a combination of pheromone signals and heuristic information. 
In this setting, the evolved heuristic defines a function that produces a heuristic matrix representing the desirability of selecting each edge. 
The function takes as input the distance matrix, node coordinates, customer demands and vehicle capacity, and outputs a matrix of the same shape indicating edge-wise preference scores. 
These heuristic values are combined with pheromone trails to guide probabilistic solution construction, enabling a balance between exploration and exploitation during the search process.

\textbf{Flow Shop Scheduling Problem (FSSP) under GLS.} We consider the flow shop scheduling problem (FSSP), where n jobs are processed on m machines in the same predetermined order. 
Each machine can process at most one job at a time, and each job must complete its operations sequentially across all machines. 
The objective is to find a job sequence that minimizes the makespan, i.e., the total completion time of all jobs. 

We adopt a Guided Local Search (GLS) framework, which operates as a perturbation-and-improvement procedure. 
Starting from a current job sequence, GLS iteratively applies local search operators, such as Swap and Relocate, to explore neighboring solutions \cite{emmons2012flow}. 
To escape local optima, GLS maintains a heuristic-guided penalty mechanism that dynamically modifies the execution-time matrix and prioritizes certain jobs or operations for perturbation. In this setting, the evolved heuristic jointly determines (i) how to update the execution-time matrix and (ii) which subset of jobs should be perturbed at each iteration. 
The function takes as input the current job sequence and the execution-time matrix, and outputs an updated matrix together with selected jobs for perturbation, thereby guiding the search toward schedules with reduced makespan.

\textbf{Seed Heuristics Used.} Our SimpleEvol can run without additional seed heuristics.
This is because the existing LLMs are knowledgeable enough to craft a simple baseline heuristic. By adding only descriptions of the intended problem and evolved heuristics, we enable the LLM to search freely and define the start of exploration at the beginning, without instilling external knowledge.
The seed heuristic used for FSSP under the GLS framework for all AHD methods is shown below.
\vspace{1em}
\renewcommand{\lstlistingname}{Seed Heuristic}

\lstdefinestyle{customcode}{
    backgroundcolor=\color{codebg},
    basicstyle=\ttfamily\small\color{codeblue},
    keywordstyle=\color{codeblue},
    commentstyle=\color{gray},
    stringstyle=\color{codeblue},
    showstringspaces=false,
    frame=single,
    rulecolor=\color{gray},
    breaklines=true,
    tabsize=2
}

\begin{lstlisting}[style=customcode,language=Python, caption={FSSP heuristic under GLS framework},captionpos=b]
import numpy as np

def get_matrix_and_jobs(current_sequence: np.ndarray, time_matrix: np.ndarray, m: int, n: int) -> tuple[np.ndarray, np.ndarray]:

    # keep matrix unchanged
    new_matrix = time_matrix.copy()

    # compute job total processing time
    job_sum = np.sum(time_matrix, axis=1)

    # select top-k jobs (EOH commonly uses k <= 5)
    k = min(3, n)
    perturb_jobs = np.argsort(-job_sum)[:k]

    return new_matrix, perturb_jobs

# === Description ===
# Trivial seed heuristic for FSSP-GLS for experiment 0 only.
# This heuristic selects jobs with the largest total processing time,
# without considering sequence structure or machine interactions.
\end{lstlisting}

\begin{lstlisting}[style=customcode,language=Python, caption={CVRP heuristic under ACO framework},captionpos=b]

import numpy as np

def heuristics(distance_matrix: np.ndarray, coordinates: np.ndarray, demands: np.ndarray, capacity: int) -> np.ndarray:

     return 1 / distance_matrix

# A simple inverse-distance seed heuristic for CVRP-ACO.
# It assigns larger heuristic values to shorter edges and ignores demand, depot structure, and residual capacity effects.
# This seed serves as a minimal distance-only baseline for experiment 0.
\end{lstlisting}

\begin{lstlisting}[style=customcode,language=Python, caption={TSP heuristic under Step-by-step Construction},captionpos=b]

import numpy as np

def select_next_node(current_node: int, destination_node: int, unvisited_nodes: set, distance_matrix: np.ndarray) -> int:
    # Select the next node to visit from the unvisited nodes.
    scores = {}
    for node in unvisited_nodes:
        scores[node] = 1
    next_node = min(scores, key=scores.get)
    return next_node

\end{lstlisting}
For completeness and reproducibility, we also report the seed heuristics used by other AHD frameworks for TSP and CVRP, though SimpleEvol does not require seed heuristics. A random selection function is used for TSP, and a simple inverse-distance heuristic is adopted for CVRP-ACO.

\section{Calculations of Proposed Metrics}
\label{sec:calculation_of_metrics}

\subsection{Calculations of AHD Complexity Score (AHI)}
\label{sec:details_of_AHI}

We provide detailed calculations of the AHD Handcraftedness Index (AHI) for the evaluated frameworks. Recall that
\[
\text{AHI}(A) = M + K + \log_{10}(1 + Q),
\]
where $M$ is the number of indispensable framework modules, $K$ is the number of distinct LLM operation types, and $Q$ is the total number of LLM calls in one run averaged across all models. For all frameworks, we compute AHI based on the actual algorithmic implementations in their official codebases, rather than the descriptions in the original papers, as the two occasionally differ.

\textbf{Initialization.} We do not include initialization (bootstrapping) as a separate operation type in $K$, since all AHD frameworks require an initial generation stage, and differences in initialization strategies (e.g., number of initial samples) are treated as implementation details rather than structural complexity.

\vspace{4pt}
\noindent
\textbf{ReEvo.}
For ReEvo \cite{ye2024reevo}, we have:
\begin{itemize}
    \item $M = 2$: (i) a single LLM for reflection and code generation, (ii) a random selection function to manage population.
    \item $K = 4$: distinct LLM operations including short-term reflection, long-term reflection, crossover, and mutation.
    \item $Q = 1293$: average number of LLM calls per run on TSP, as reported in Table~\ref{tab:aci_comparison}.
\end{itemize}
Thus, $\text{AHI} = 2 + 4 + \log_{10}(1 + 1293) = 9.112$.

\vspace{4pt}
\noindent
\textbf{FunSearch.}
For FunSearch \cite{romera2024mathematical}, we have:
\begin{itemize}
    \item $M = 3$: (i) a code generator LLM; 
    (ii) an intra-island structure module, which maintains clusters of programs and samples previous functions within each island to construct prompts; 
    and (iii) an inter-island controller, which maintains multiple islands and periodically resets weak islands using programs from stronger islands.
    \item $K = 1$: a single LLM operation type corresponding to prompt-conditioned generation. After bootstrapping, the prompt contains previous heuristic versions sampled from the database, and the LLM is asked to complete a new improved function.
    \item $Q = 821$: average number of LLM calls per run, as reported in Table~\ref{tab:aci_comparison}.
\end{itemize}
Thus, $\text{AHI} = 3 + 1 + \log_{10}(1 + 821) = 6.915$. 

\vspace{4pt}
\noindent
\textbf{MCTS-AHD.}
For MCTS-AHD \cite{zheng2025monte}, we have:
\begin{itemize}
    \item $M = 2$: 
    (i) an LLM-based heuristic generator that produces new heuristic functions and their descriptions, and 
    (ii) an MCTS controller that maintains the search tree and determines which heuristic states are selected and expanded.
    
    \item $K = 5$: five distinct LLM search operations excluding initialization, including 
    two mutation actions (\texttt{m1}, \texttt{m2}), 
    two crossover actions (\texttt{e1}, \texttt{e2}), 
    and one tree-path reasoning action (\texttt{s1}).
    
    \item $Q = 1640$: average total number of LLM calls per run on TSP.
\end{itemize}
Thus, $\text{AHI} = 2 + 5 + \log_{10}(1 + 1640) = 10.215$.

\vspace{4pt}
\noindent
\textbf{SimpleEvol.}
For our SimpleEvol framework, we have:
\begin{itemize}
    \item $M = 1$: (i) a single LLM for code generation and summarization.
    \item $K = 2$: two LLM operations including feedback-conditioned code generation and context compression (summary).
    \item $Q = 1001$: average number of LLM calls per run, as reported in Table~\ref{tab:aci_comparison}.
\end{itemize}
Thus, $\text{AHI} = 1 + 2 + \log_{10}(1 + 1001) = 6.001$. Note that LLM-driven sub-processes such as history compression are counted under K rather than M, since M counts structural components and K counts distinct LLM invocation types—counting an LLM-only sub-process in both would double-count the same complexity.

\paragraph{Sensitivity to AHI weights.}
The default AHI uses an unweighted additive form to avoid introducing additional hyperparameters. To examine whether the observed AHI--ICE relationship depends on this equal-weight specification, we further consider
\[
\mathrm{AHI}_w
=
w_M M
+
w_K K
+
w_Q \log_{10}(1+Q),
\]
where
$w_M,w_K,w_Q \in \{0.5,0.75,1,1.25,1.5,2\}$.
We exhaustively evaluate all $6^3=216$ weight combinations. SimpleEvol remains the lowest-AHI framework in 174 of the 216 settings (80.56\%), while the original AHI ranking is preserved in 168 settings (77.78\%). More importantly, the association between AHI and ICE remains negative under every tested weighting combination on both TSP and CVRP. Across all settings, the median Spearman correlation is $-1.0$, with a mean of approximately $-0.944$. These results indicate that the observed inverse AHI--ICE relationship is not specific to the equal-weight formulation used in the main analysis.

\vspace{4pt}
\noindent

\subsection{Details of Model Intelligence Score $I(m)$}
\label{sec:details_I(m)}

\paragraph{Benchmark selection.}
We select four benchmarks that collectively cover the capability dimensions most directly relevant to LLM-based AHD: broad knowledge and reasoning (MMLU-Pro~\cite{wang2024mmlu}), instruction following (IFBench~\cite{pyatkin2025ifbench}), mathematical reasoning (AIME 2025~\cite{ye2025aime}) and code generation (LiveCodeBench~\cite{jain2024livecodebench}). This selection is motivated by the nature of the AHD task itself: effective heuristic design requires a model to have sufficient optimization background knowledge to understand heterogeneous tasks and instructions, adhere to the requirements of demanding operations (e.g., how to effectively compare and recombine two complicated parents), mathematically reason about combinatorial structures and algorithmic trade-offs, and translate ideas into correct, executable implementations. A model that is strong across all four dimensions is thus more likely to consistently devise and refine heuristics across diverse problem settings. Crucially, these four dimensions are not perfectly correlated, as shown in Table~\ref{tab:intelligence_scores}, in which models that score similarly on one benchmark can differ substantially on others (e.g., o3-mini and Qwen3-235B-Thinking have similar MMLU-Pro scores but diverge significantly on IFBench), suggesting that each benchmark captures a genuinely distinct aspect of capability. 



\begin{table*}[h]
\centering
\setlength{\tabcolsep}{4.5pt}
\renewcommand{\arraystretch}{0.85}
\begin{tabular}{lcc cc cc cc c}
\toprule
 & \multicolumn{8}{c}{Benchmark} & \\
\cmidrule(lr){2-9}
Model 
& \multicolumn{2}{c}{MMLU-Pro} 
& \multicolumn{2}{c}{IFBench} 
& \multicolumn{2}{c}{AIME 2025} 
& \multicolumn{2}{c}{LiveCodeBench} 
& $I(m)$ \\
\cmidrule(lr){2-3} \cmidrule(lr){4-5} \cmidrule(lr){6-7} \cmidrule(lr){8-9}
 & Score & Ratio & Score & Ratio & Score & Ratio & Score & Ratio & \\
\midrule

GPT-4o-mini & 64.8 & 1.00 & 31.0 & 1.00 & 14.7 & 1.00 & 23.4 & 1.00 & 1.00 \\
GPT-4.1-nano & 65.7 & 1.01 & 32.0 & 1.03 & 24.0 & 1.63 & 32.6 & 1.39 & 1.24 \\
Claude Sonnet 3.7 & 80.3 & 1.27 & 44.0 & 1.42 & 21.0 & 1.43 & 39.4 & 1.68 & 1.44 \\
DeepSeek-v3 & 81.9 & 1.30 & 41.0 & 1.32 & 41.0 & 2.79 & 40.5 & 1.73 & 1.70 \\
GPT-4.1-mini & 78.1 & 1.24 & 38.3 & 1.24 & 46.3 & 3.15 & 48.3 & 2.06 & 1.77 \\
Gemini-2.5-Flash & 80.9 & 1.28 & 39.0 & 1.26 & 60.3 & 4.10 & 49.5 & 2.12 & 1.93 \\
Qwen3-235B-Instruct & 82.8 & 1.31 & 46.1 & 1.49 & 71.7 & 4.88 & 52.4 & 2.24 & 2.15 \\
o3-mini  & 79.1 & 1.25 & 67.1 & 2.16 & 75.6 & 5.14 & 71.7 & 3.06 & 2.56 \\
Qwen3-235B-Thinking & 84.3 & 1.33 & 51.2 & 1.65 & 91.0 & 6.19 & 78.8 & 3.37 & 2.60 \\
GPT-5-mini & 82.8 & 1.31 & 71.2 & 2.30 & 85.0 & 5.78 & 69.2 & 2.96 & \textbf{2.68} \\
\bottomrule
\end{tabular}

\caption{
Benchmark scores (Score) and normalized ratios (Ratio) used to compute the model intelligence score $I(m)$. Ratios are normalized with respect to the weakest model (GPT-4o-mini), and $I(m)$ is computed based on aggregated normalized performance across benchmarks.
}
\label{tab:intelligence_scores}
\end{table*}



\paragraph{Interpretation.}
$I(m)$ is a relative proxy for model intelligence related to the AHD task, not an all-encompassing measure of overall model intelligence. Its purpose is to provide a unified, framework-agnostic scalar that enables systematic comparison of how different AHD frameworks scale with model intelligence. Scores are collected from the Artificial Analysis evaluation platform~\cite{artificialanalysis2026} under standardized settings.

\textbf{Benchmark coverage.} The capability dimensions captured by each benchmark are described as:

\begin{itemize}
    \item \textbf{MMLU-Pro} measures broad multi-domain knowledge and general reasoning ability across diverse academic subjects, reflecting the model’s capability to understand heterogeneous problem contexts.
    \item \textbf{IFBench} evaluates instruction-following under complex and compositional task requirements, capturing the ability to adhere to detailed specifications in heuristic design and refinement.
    \item \textbf{AIME 2025} measures advanced mathematical reasoning, particularly structured problem solving and multi-step symbolic derivation.
    \item \textbf{LiveCodeBench} evaluates practical code generation ability in realistic programming tasks, capturing the model’s ability to translate ideas into executable implementations, debug and repair, and predict their execution outcomes.
\end{itemize}



\clearpage

\section{Extended Experimental Results}
\label{sec:more_experiments_results}

\subsection{Heuristics Evolved by SimpleEvol}
We further analyze the heuristics evolved by \textbf{SimpleEvol} under different LLM backbones. 
We focus on a controlled setting where 
the \emph{same framework} is used with models of varying intelligence levels. 
This allows us to isolate the effect of model intelligence on the quality of evolved heuristics. The performance of heuristics is evaluated by the fitness score (i.e., the objective score or $g(h)$) on the training dataset. A heuristic is better with a lower fitness score.

Specifically, we compare the best heuristics discovered by \texttt{GPT-4.1-nano} and 
\texttt{GPT-5-mini}. According to our intelligence metric $I(m)$, \texttt{GPT-5-mini} 
has approximately twice the intelligence score of \texttt{GPT-4.1-nano}. 
Despite using the same SimpleEvol framework and identical evaluation budget, 
the heuristic evolved by \texttt{GPT-5-mini} achieves substantially better performance.

More importantly, the improvement is not merely in objective value, but also in 
\emph{structural sophistication}. The stronger model is able to synthesize more 
advanced heuristic components, such as insertion-based structural reasoning, 
scale-normalized regret, and topology-aware priors, while the weaker model tends 
to rely on simpler greedy rollout strategies.

These observations suggest that, under a fixed AHD framework, heuristic quality 
can scale significantly with model intelligence. This provides empirical evidence 
that the primary bottleneck in AHD may be shifting from framework-level complexity 
to model-level intelligence, consistent with the scaling behavior observed in 
modern LLMs.


\renewcommand{\lstlistingname}{Evolved Heuristic}

\lstset{
    basicstyle=\ttfamily\footnotesize,
    breaklines=true,
    frame=single,
    framerule=0.6pt,
    rulecolor=\color{gray!45},
    backgroundcolor=\color{gray!5},
    keywordstyle=\color{blue!70!black},
    commentstyle=\color{green!50!black},
    stringstyle=\color{red!70!black},
    showstringspaces=false,
    columns=fullflexible,
}

\begin{lstlisting}[language=Python,caption={TSP constructive heuristic by GPT-4.1-nano; Fitness score: 6.15},captionpos=b]
"""
Weighted nearest-neighbor rollout. For each candidate next node, estimate full completion cost
by greedy nearest-neighbor chaining to destination. Final score = alpha * immediate distance
+ beta * rollout estimate, with stronger early emphasis on immediate distance.
"""

def select_next_node(current_node: int, destination_node: int, unvisited_nodes: set, distance_matrix: np.ndarray) -> int:
    import numpy as np
    unvisited_list = list(unvisited_nodes)
    if not unvisited_list:
        return destination_node

    # Greedy rollout estimate from a candidate node to destination
    def evaluate_node(node):
        dist_curr = distance_matrix[current_node, node]
        remaining = unvisited_list.copy()
        remaining.remove(node)
        total_est = dist_curr
        current = node
        while remaining:
            next_node = min(remaining, key=lambda n: distance_matrix[current, n])
            total_est += distance_matrix[current, next_node]
            current = next_node
            remaining.remove(next_node)
        total_est += distance_matrix[current, destination_node]
        return total_est

    # Dynamic weighting: more aggressive local bias in early stage
    progress_ratio = 1 - len(unvisited_list) / distance_matrix.shape[0]
    alpha = 0.75 * (1 - progress_ratio) + 0.25 * progress_ratio
    beta = 1 - alpha

    # Combine immediate distance and rollout estimate
    weighted_scores = [alpha * distance_matrix[current_node, n] + beta * evaluate_node(n) for n in unvisited_list]
    best_idx = np.argmin(weighted_scores)
    return unvisited_list[best_idx]
\end{lstlisting}

\begin{lstlisting}[language=Python,caption={TSP constructive heuristic by GPT-5-mini; Fitness score: 5.97},captionpos=b]
"""
Fast prefiltered adaptive insertion. One-shot MST on {current, unvisited, destination} to obtain node degrees and two-edge lower bounds. Candidate prefilter via proxy: dist(curr,u) + dist(u,dest) + two-edge sum - MST-degree (top-K). For each candidate: vectorized insertion deltas, normalized regret (second-best minus best, std-normalized), plus small sampled lookahead (L <= 4 ~ sqrt(remaining)) using minimal insertion costs of nearest nodes. Final selection via lexicographic tuple (regret, mst_deg, -two_sum, -future, -best_cost, proximity, return_prox, -id). Key params: K >= 6, L <= 4.
"""

import numpy as np

def select_next_node(current_node:int,destination_node:int,unvisited_nodes:set,distance_matrix:np.ndarray)->int:
    curr=int(current_node); dest=int(destination_node)
    unvisited=[int(x) for x in unvisited_nodes]

    # trivial cases
    if not unvisited: return int(dest)
    if len(unvisited)==1: return int(unvisited[0])
    dist=np.asarray(distance_matrix,float)
    tour=[curr,dest]; remaining=set(unvisited)
    # build node set
    S_nodes=[curr]+[u for u in unvisited]+[dest]
    S_nodes=list(dict.fromkeys(S_nodes))
    idx_of={node:i for i,node in enumerate(S_nodes)}
    nS=len(S_nodes)
    # distance submatrix
    distS=np.empty((nS,nS),dtype=float)
    for i in range(nS): distS[i,:]=dist[S_nodes[i],S_nodes]
    # MST (Prim)
    INF=1e18
    in_mst=np.zeros(nS,dtype=bool)
    key=np.full(nS,INF,dtype=float)
    parent=np.full(nS,-1,dtype=int)
    key[0]=0.0
    degrees=np.zeros(nS,dtype=int)

    for _ in range(nS):
        mask=~in_mst
        if not np.any(mask): break
        cand=np.where(mask)[0]
        u=int(cand[np.argmin(key[mask])])
        if in_mst[u]: break
        in_mst[u]=True; p=parent[u]
        if p!=-1: degrees[u]+=1; degrees[p]+=1
        cand_idx=np.nonzero(~in_mst)[0]
        vals=distS[u,cand_idx]
        improved=vals<key[cand_idx]
        if np.any(improved):
            key[cand_idx[improved]]=vals[improved]
            parent[cand_idx[improved]]=u

    # two-smallest-edge sum
    two_small_sum={}
    for i,node in enumerate(S_nodes):
        if nS<=1: two_small_sum[node]=0.0; continue
        row=np.delete(distS[i],i)
        k=min(2,row.size)
        if k==0: two_small_sum[node]=0.0
        else:
            part=np.partition(row,k-1)[:k]
            two_small_sum[node]=float(np.sum(part))

    # main loop
    while remaining:
        rem_list=list(remaining); nrem=len(rem_list)
        # candidate prefilter
        K=min(nrem,max(6,int(max(6,2*np.sqrt(nrem)))))
        proxies=[]
        for u in rem_list:
            mst_deg=int(degrees[idx_of[u]]) if u in idx_of else 0
            two_sum=float(two_small_sum.get(u,0.0))
            proxy=dist[curr,u]+dist[u,dest]+0.3*two_sum-0.5*mst_deg
            proxies.append((proxy,u))
        proxies.sort(key=lambda x:x[0])
        candidates=[p[1] for p in proxies[:K]]
        edges_a=np.array(tour[:-1],dtype=int)
        edges_b=np.array(tour[1:],dtype=int)
        edge_base=dist[edges_a,edges_b] if edges_a.size>0 else np.array([],dtype=float)
        best_choice=None; best_score=None; best_insert_pos=None
        L=min(4,max(1,int(np.sqrt(max(1,nrem)))))

        for u in candidates:
            # insertion cost
            if edges_a.size==0: cost_vec=np.array([0.0])
            else: cost_vec=dist[edges_a,u]+dist[u,edges_b]-edge_base
            if cost_vec.size==0: continue
            sorted_idx=np.argsort(cost_vec)
            best_cost=float(cost_vec[sorted_idx[0]])
            second_cost=float(cost_vec[sorted_idx[1]]) if cost_vec.size>1 else best_cost
            std_costs=float(np.std(cost_vec)) if cost_vec.size>1 else 0.0
            normalized_regret=(second_cost-best_cost)/(1.0+std_costs)
            insert_pos=int(sorted_idx[0])+1

            # sampled lookahead
            future_sum=0.0
            others=[x for x in remaining if x!=u]
            if others:
                row=dist[u,np.array(others,dtype=int)]
                take=min(L,row.size)
                if take>0:
                    idxs=np.argpartition(row,take-1)[:take]
                    sample=np.array(others,dtype=int)[idxs]
                    tour2=tour.copy(); tour2.insert(insert_pos,u)
                    a2=np.array(tour2[:-1],dtype=int)
                    b2=np.array(tour2[1:],dtype=int)
                    base2=dist[a2,b2] if a2.size>0 else np.array([],dtype=float)
                    for v in sample:
                        costs_v=dist[a2,v]+dist[v,b2]-base2
                        if costs_v.size>0: future_sum+=float(np.min(costs_v))
            mst_deg=int(degrees[idx_of[u]]) if u in idx_of else 0
            twosum=float(two_small_sum.get(u,0.0))

            # lexicographic score
            score=(normalized_regret,mst_deg,-twosum,-future_sum,-best_cost,
                   -dist[curr,u],-dist[u,dest],-u)
            if best_score is None or score>best_score:
                best_score=score; best_choice=u; best_insert_pos=insert_pos
        if best_choice is None: break
        tour.insert(best_insert_pos,int(best_choice)); remaining.remove(best_choice)

    return int(tour[1]) if len(tour)>=2 else int(dest)
\end{lstlisting}

\newpage
\subsection{Results on TSPLib}
\label{sec:result_on_tsplib}
We further evaluate the best heuristics for TSP-constructive discovered by SimpleEvol and other baselines under \texttt{GPT-4.1-nano} on a real-world benchmark, namely TSPLIB~\cite{reinelt1991tsplib} instances, to assess their out-of-distribution generalization. 
As shown in Table~\ref{tab:tsplib_gpt5mini}, SimpleEvol achieves the lowest average optimality gap, obtaining the best average gap (10.26\%) among all compared AHD methods. 
More importantly, SimpleEvol attains the largest number of top-1 results across 15 instances (\#top1 = 8), indicating strong and broad instance-level competitiveness on this out-of-distribution benchmark.


\begin{table}[h]
\centering
\caption{Results of LLM-based AHD methods for TSP on TSPLIB instances. Following the same protocol as ReEvo~\cite{ye2024reevo}, we compute the reported optimality gap from the average over three runs with different starting nodes. For each instance, the best result across all methods is marked in \textbf{bold}.}
\label{tab:tsplib_gpt5mini}
\small
\setlength{\tabcolsep}{18pt}
\renewcommand{\arraystretch}{1.12}
\begin{tabular}{l|cccc}
\thickhline
\thickhline
Instance & EoH & ReEvo & FunSearch & SimpleEvol (ours) \\
\hline
bier127.tsp & 13.60\% & 8.33\% & 20.69\% & \textbf{7.16\%} \\
ch130.tsp   & 10.63\% & 12.24\% & 20.04\% & \textbf{10.12\%} \\
eil51.tsp   & 8.85\%  & 14.20\% & \textbf{7.27\%} & 11.96\% \\
fl417.tsp   & 13.49\% & 10.59\% & 33.06\% & \textbf{10.50\%} \\
kroA150.tsp & \textbf{8.44\%} & 9.86\% & 25.48\% & 10.36\% \\
kroB100.tsp & 11.57\% & 10.28\% & 12.90\% & \textbf{9.93\%} \\
kroC100.tsp & 11.47\% & \textbf{8.16\%} & 14.18\% & 11.16\% \\
lin318.tsp  & 12.34\% & 10.95\% & 27.57\% & \textbf{10.11\%} \\
pr226.tsp   & 15.62\% & 10.11\% & 28.34\% & \textbf{8.86\%} \\
pr264.tsp   & 12.36\% & \textbf{10.97\%} & 14.30\% & 11.48\% \\
pr299.tsp   & \textbf{10.11\%} & 15.06\% & 33.35\% & 11.46\% \\
d493.tsp    & 15.50\% & 13.15\% & 31.44\% & \textbf{12.17\%} \\
rat99.tsp   & \textbf{8.85\%} & 9.18\% & 14.71\% & 10.19\% \\
ts225.tsp   & \textbf{5.60\%} & 10.33\% & 10.04\% & 6.26\% \\
pr439.tsp   & 13.24\% & 15.91\% & 30.23\% & \textbf{12.15\%} \\
\hline
AVG     & 11.44\% & 11.29\% & 21.57\% & \textbf{10.26\%} \\
\#top1  & 4 & 2 & 1 & 8 \\
\thickhline
\thickhline
\end{tabular}
\end{table}

\subsection{Robustness of AHI--ICE relationship}
\label{sec:ICE_variants}

While the slope of the regression line serves as the primary metric to evaluate ICE in this work, it alone does not explicitly reflect the sensitivity to the presence of individual models nor the benchmark set $\mathcal{B}$ selection used to compute intelligence score $I(m)$. Regression-based ICE captures the global trend between model intelligence and optimization performance across the full set of evaluated models, while a leave-one-out analysis provides a complementary perspective by characterizing the stability of this trend with respect to the model set composition. We therefore report leave-one-out ICE for model-selection sensitivity and additionally recompute ICE using an independent intelligence index to assess benchmark-selection sensitivity.

\paragraph{Leave-One-Out Regression ICE.}
We evaluate the stability of regression-based ICE by recomputing the slope after removing each model. Formally, for each model $i$, we define:
\begin{equation}
\text{ICE}_{\text{LOO}}^{(i)}(\mathcal{A}) = \alpha_\mathcal{A}^{(-i)},
\end{equation}
where $\alpha_{\mathcal{A}}^{(-i)}$ is the regression slope computed without the $i$-th model. We report the mean and standard deviation across all leave-one-out estimates:
\begin{equation}
\text{ICE}_{\text{LOO-mean}}(\mathcal{A}), \quad \text{ICE}_{\text{LOO-std}}(\mathcal{A}).
\end{equation}

\begin{table}[t]
\centering
\caption{Robustness of AHI--ICE relationship under leave-one-out regression and an alternative intelligence metric. 
$\mathrm{ICE}_{\mathrm{AA}}$ is computed by replacing our benchmark-based $I(m)$ with the Artificial Analysis Intelligence Index~\cite{artificialanalysis2026index}. The highest value of ICE for each setting is marked in bold, while the second largest value is underlined.}
\label{tab:ice_loo_robustness}
\small
\setlength{\tabcolsep}{5.5pt}
\renewcommand{\arraystretch}{1.12}
\begin{tabular}{l|cccc|cccc}
\thickhline
\thickhline
\multirow{2}{*}{Method}
& \multicolumn{4}{c|}{TSP Constructive}
& \multicolumn{4}{c}{CVRP-ACO} \\
\cline{2-9}
& ICE & LOO-mean & LOO-std & $\mathrm{ICE}_{\mathrm{AA}}$
& ICE & LOO-mean & LOO-std & $\mathrm{ICE}_{\mathrm{AA}}$ \\
\hline
FunSearch  
& \underline{1.8174} & \underline{1.8110} & 0.4258 & \underline{2.2103}
& \underline{5.4381} & \underline{5.4112} & 0.9423 & \underline{6.8866} \\
EoH        
& 1.5082 & 1.4969 & 0.4040 & 1.8132
& 3.6572 & 3.6853 & 0.9137 & 5.5700 \\
ReEvo      
& 0.8230 & 0.8201 & 0.2981 & 1.4627
& 2.0824 & 2.0223 & 1.1802 & 5.3194 \\
MCTS-AHD
& 0.7525 & 0.7560 & 0.3087 & 1.2131
& 1.5380 & 1.5351 & 0.8073 & 3.9170 \\
SimpleEvol 
& \textbf{2.1941} & \textbf{2.1903} & 0.5558 & \textbf{2.5764}
& \textbf{6.2128} & \textbf{6.2147} & 1.1583 & \textbf{7.1379} \\
\thickhline
\thickhline
\end{tabular}
\end{table}

\paragraph{Extension to MCTS-AHD.}
To broaden the framework-level evaluation beyond the four methods compared in the main text,
we additionally evaluate MCTS-AHD~\cite{zheng2025monte}, which adopts a structurally distinct
tree-search paradigm. Following the same AHI counting rules, MCTS-AHD obtains an AHI of
10.215, while its ICE values are 0.7525 on TSP Constructive and 1.5380 on CVRP-ACO.
Its inclusion preserves the same inverse ordering between framework complexity and ICE on both
tasks, providing additional evidence that the observed AHI--ICE pattern is not specific to the
four main compared frameworks.

\paragraph{Model Selection Sensitivity.}
Table~\ref{tab:ice_loo_robustness} shows that the regression-based ICE estimates are stable under leave-one-out perturbations of the model set. On both TSP Constructive and CVRP-ACO, the LOO-mean values closely match the original regression ICE for all five frameworks, with deviations within $\pm 0.07$. This indicates that the estimated ICE values are not dominated by any single backbone model. More importantly, the relative ordering of frameworks is preserved under LOO on both tasks. SimpleEvol remains the highest-ICE framework, followed by FunSearch, EoH, ReEvo, and MCTS-AHD. This ordering is consistent with the AHI ranking, where SimpleEvol and FunSearch lie in the
lower-complexity regime, EoH and ReEvo introduce more human-designed search components, and
MCTS-AHD has the highest AHI among the evaluated frameworks. The LOO results therefore support the same conclusion as the main regression analysis, showing that lower-complexity frameworks tend to exhibit higher intelligence conversion efficiency.

The separation is especially clear between the lighter and more heavily orchestrated frameworks.
SimpleEvol maintains a substantial ICE margin over EoH, ReEvo, and MCTS-AHD on both tasks,
while FunSearch also consistently exceeds ReEvo and MCTS-AHD. The difference between SimpleEvol and FunSearch is relatively moderate, which is expected given that both frameworks have relatively low and close AHI scores. This suggests that the main empirical pattern is not merely that SimpleEvol outperforms every baseline by a large margin, but that frameworks with lower structural complexity form a higher-ICE group than more heavily engineered evolutionary pipelines. Overall, the five-framework comparison strengthens the empirical pattern that
lighter handcrafted orchestration is associated with higher ICE.

\paragraph{Alternative intelligence metric.}
To examine whether the ICE ranking is sensitive to our construction of $I(m)$, we further recompute ICE using an independent composite intelligence score from Artificial Analysis, denoted as $I_{\mathrm{AA}}(m)$. This index aggregates ten challenging evaluations across mathematics, science, coding, and reasoning, and is constructed independently from our downstream AHD experiments. Replacing $I(m)$ with $I_{\mathrm{AA}}(m)$ preserves the same ICE ordering
shown in Table~\ref{tab:ice_loo_robustness}:
SimpleEvol $>$ FunSearch $>$ EoH $>$ ReEvo $>$ MCTS-AHD. This suggests that the observed AHI--ICE pattern is not an artifact of our particular benchmark selection for model intelligence. 

\subsubsection{Statistical Uncertainty of ICE Differences}
\label{sec:ice_statistical_evidence}

Beyond the point estimates of ICE, we further quantify the uncertainty in pairwise differences between SimpleEvol and the compared frameworks using the Bayesian bootstrap~\cite{rubin1981bayesian}. By repeatedly reweighting the evaluated models with Dirichlet weights and refitting the regressions, it captures the uncertainty of ICE differences induced by model-level variability.  
Specifically, for each of 10,000 bootstrap replicates, we draw a shared set of Dirichlet weights over the ten backbone models. The same weights are applied to all frameworks, and the weighted regression in Eq.~\ref{eq:ice_regression} is refitted to obtain the corresponding ICE slope. For each baseline $A$, we then compute the pairwise slope difference
\[
\Delta_A =
\mathrm{ICE}_{\mathrm{SimpleEvol}}
-
\mathrm{ICE}_{A}.
\]
Because our hypothesis concerns whether SimpleEvol exhibits a larger intelligence conversion slope, we report the posterior probability
$\Pr(\Delta_A > 0)$. As shown in Table~\ref{tab:ice_bootstrap}, all eight task--framework comparisons favor SimpleEvol directionally, with $\Pr(\Delta_A>0)$ ranging from 79.1\% to 99.9\%. Five of the eight comparisons exceed 0.95.
The evidence is particularly strong relative to ReEvo and MCTS-AHD, reaching 97.6\% and 98.9\%
on TSP and 99.7\% and 99.9\% on CVRP, respectively. The difference relative to EoH also exceeds
0.95 on TSP. These results provide additional support that the higher ICE of SimpleEvol is not
limited to the point estimates.

\begin{table}[t]
\centering
\caption{
Bayesian-bootstrap uncertainty of pairwise ICE differences.
$\Delta$ denotes
$\mathrm{ICE}_{\mathrm{SimpleEvol}}-\mathrm{ICE}_{A}$.
$\Pr(\Delta>0)$ is estimated from 10,000 Bayesian-bootstrap replicates.
}
\label{tab:ice_bootstrap}
\begin{tabular}{llcc}
\toprule
Task & Comparison & $\Delta$ & $\Pr(\Delta>0)$ \\
\midrule
TSP Constructive
& SimpleEvol $-$ FunSearch
& 0.380 & 85.5\% \\
& SimpleEvol $-$ EoH
& 0.688 & \textbf{95.3\%} \\
& SimpleEvol $-$ ReEvo
& 1.369 & \textbf{97.6\%} \\
& SimpleEvol $-$ MCTS-AHD
& 1.442 & \textbf{98.9\%} \\
\midrule
CVRP-ACO
& SimpleEvol $-$ FunSearch
& 0.771 & 79.1\% \\
& SimpleEvol $-$ EoH
& 2.562 & 88.7\% \\
& SimpleEvol $-$ ReEvo
& 4.130 & \textbf{99.7\%} \\
& SimpleEvol $-$ MCTS-AHD
& 4.675 & \textbf{99.9\%} \\
\bottomrule
\end{tabular}
\end{table}

A leave-two-model-out analysis further tests whether the observed ordering is driven by a small number of influential backbone models. Among the $\binom{10}{2}=45$ possible subsets obtained by removing two models, the ICE differences between SimpleEvol and other frameworks remain positive in at least 39 of the 45 subsets for every baseline comparison. Together with the Bayesian-bootstrap results, this indicates that the higher ICE of SimpleEvol is robust to the model set composition and persists after extending the analysis to MCTS-AHD.

\subsection{Search Dynamics and Budget-dependent ICE}
\label{sec:budget_ice}

\begin{figure}[b]
    \centering
    \begin{minipage}{0.49\linewidth}
        \centering
        \includegraphics[width=\linewidth]{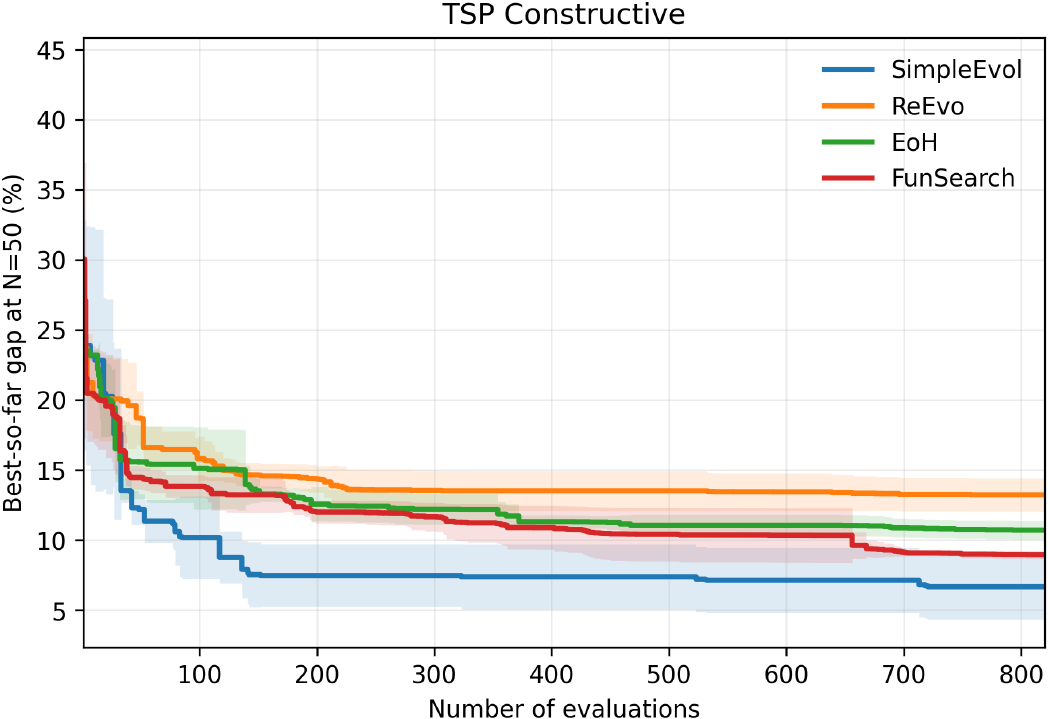}
        \small (a) TSP Constructive
    \end{minipage}
    \hfill
    \begin{minipage}{0.49\linewidth}
        \centering
        \includegraphics[width=\linewidth]{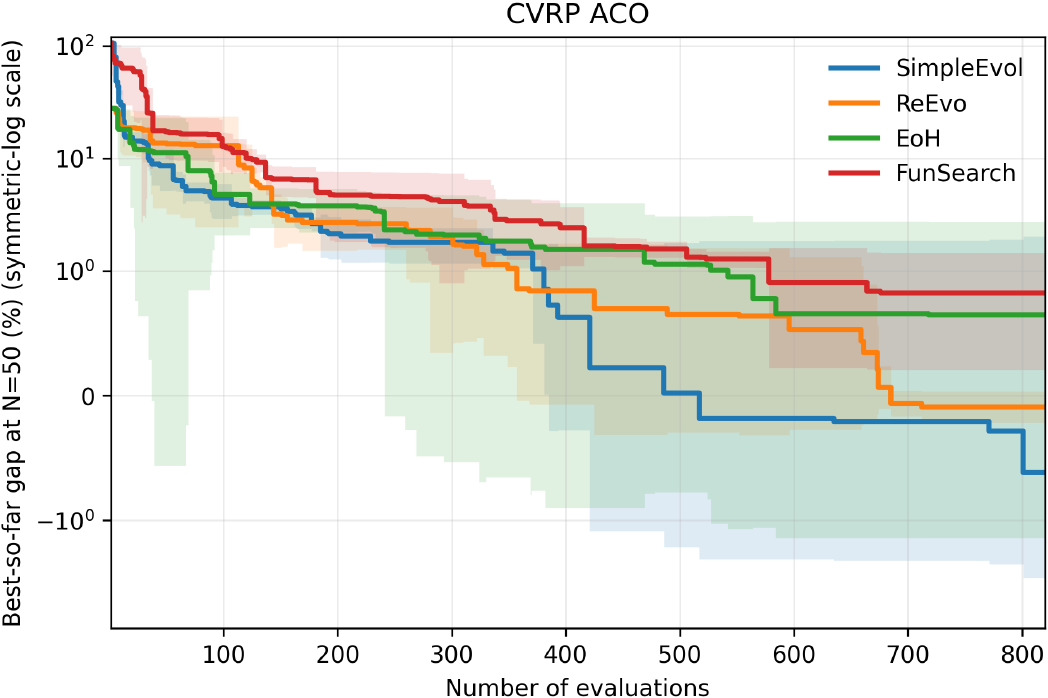}
        \small (b) CVRP-ACO
    \end{minipage}
    \caption{Best-so-far search trajectories under \texttt{GPT-5-mini}. 
    Lower gap is better. Solid curves show the mean across independent runs, 
    and shaded regions indicate $\pm$1 standard deviation. 
    Only the CVRP-ACO panel uses a symmetric-log scale to accommodate the wide range of early-stage gaps and negative gaps relative to the reference solution.}
    \label{fig:best_so_far_trajectory}
\end{figure}

The main experiments compare AHD frameworks under a common terminal budget of 820 heuristic evaluations. To examine whether the observed advantage is specific to this endpoint, we further analyze the search dynamics throughout the evaluation process. Figure~\ref{fig:best_so_far_trajectory} reports the best-so-far training gap under the same \texttt{GPT-5-mini} backbone for TSP Constructive and CVRP-ACO. The solid curves show the average best-so-far performance across independent runs, while the shaded regions indicate the corresponding variability at the run level.

On both tasks, SimpleEvol establishes competitive performance early in the search and maintains the lowest best-so-far gap over the later stages. On TSP Constructive, its advantage emerges particularly early and remains stable throughout most of the evaluation budget. On CVRP-ACO, the compared methods improve more gradually, while SimpleEvol continues to make improvements toward the end of the search. These trajectories show that SimpleEvol's advantage is sustained across multiple stages of the search and is not specific to the terminal budget of 820 evaluations.

Table~\ref{tab:budget_ice} further examines how intelligence conversion evolves with the search budget. We recompute ICE independently at several intermediate evaluation checkpoints using the best heuristic available at each checkpoint for every backbone model.

\begin{table}[t]
\centering
\caption{ICE at different heuristic-evaluation budgets. The highest ICE at each checkpoint within each task is shown in bold.}
\label{tab:budget_ice}

\setlength{\tabcolsep}{5pt}

\begin{tabular}{llccccc}
\toprule
Task & Method & 50 & 100 & 200 & 400 & 600 \\
\midrule
TSP Constructive
& ReEvo      & 0.585 & 0.783 & 0.891 & 0.981 & 0.843 \\
& EoH        & \textbf{1.393} & 0.568 & 1.046 & 1.290 & 1.582 \\
& FunSearch  & 0.482 & \textbf{0.862} & 1.328 & 1.315 & 1.776 \\
& SimpleEvol & 0.936 & 0.823 & \textbf{1.403} & \textbf{1.835} & \textbf{2.436} \\
\midrule
CVRP-ACO
& ReEvo      & 1.012 & 1.158 & 1.420 & 1.892 & 2.225 \\
& EoH        & 0.213 & 1.318 & 2.868 & 2.560 & 3.215 \\
& FunSearch  & \textbf{1.153} & \textbf{2.430} & 2.195 & 3.249 & 4.912 \\
& SimpleEvol & 0.513 & 2.215 & \textbf{3.281} & \textbf{4.028} & \textbf{6.137} \\
\bottomrule
\end{tabular}
\end{table}

At very small budgets, the ICE ordering is less stable, which is expected because only a limited portion of the  search trajectory has been observed and search performance is more strongly influenced by initialization variability. As the budget increases, however, a clearer pattern emerges. SimpleEvol achieves the highest ICE on both tasks from 200 evaluations onward, and its advantage generally becomes more pronounced at larger budgets. This suggests that the higher ICE reported at the terminal budget is not specific to the choice of 820 evaluations, but emerges progressively as stronger models obtain sufficient opportunity to influence the heuristic search.


\subsection{Extension to FSSP under the GLS Framework}
\label{sec:extension_to_fssp_gls}

\begin{figure}[h]
    \centering
    \includegraphics[width=\linewidth]{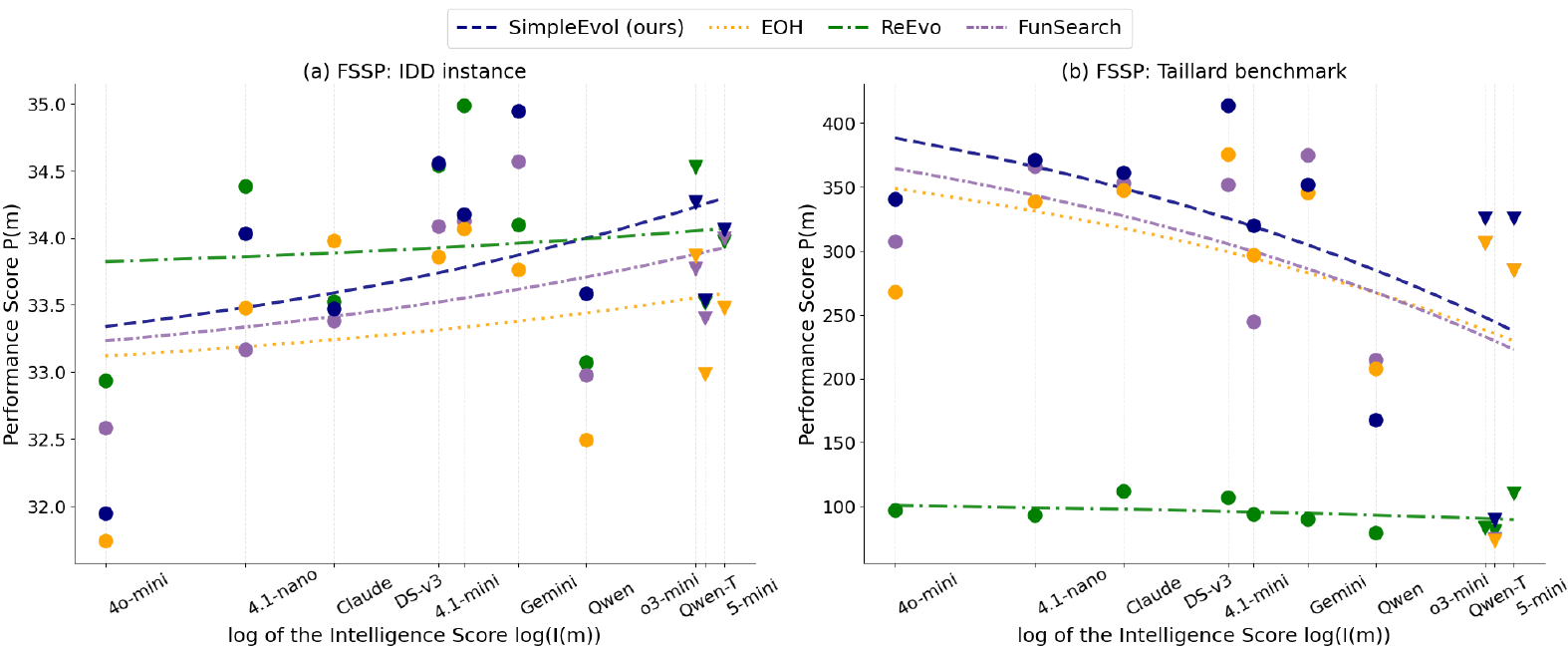}
    \caption{Relationship between model intelligence and AHD performance on FSSP-GLS. Left: 64-instance held-out IDD synthetic instances. Right: Taillard benchmark instances.}
    \label{fig:fssp_pi_plot}
\end{figure}

To examine whether our observations extend beyond routing-style problems, we further evaluate the four AHD frameworks on the FSSP under the GLS framework. In Figure~\ref{fig:fssp_pi_plot}, we report the relationship between model intelligence and the performance score on the held-out IDD dataset consisting of 64 instances, and study the generalization performance of the evolved heuristics on the classical Taillard benchmark set \cite{taillard1993benchmarks}. The details of average gaps and performance metric values of each (method, model) are in Table \ref{tab:idd_ood_comparison_fssp}.

\paragraph{In-distribution improvement.}
On the IDD synthetic distribution, stronger backbone models generally produce better heuristics, and the fitted trends are positive across all compared frameworks. More importantly, SimpleEvol still leads in ICE (i.e., the slope of the regression line), and the ICE of the other baselines broadly decreases as their AHI increases. This suggests that simpler frameworks also tend to convert model intelligence more efficiently in the FSSP-GLS training distribution.

\paragraph{Out-of-distribution generalization.}
The pattern changes on the Taillard benchmark. Increasing model intelligence no longer consistently improves performance; instead, the fitted slopes become negative across the compared frameworks. This suggests that stronger models may optimize more effectively for the synthetic training distribution while also exploiting distribution-specific regularities that do not transfer equivalently to heterogeneous benchmark instances. However, this negative trend does not imply poor OOD performance of SimpleEvol. In absolute terms, SimpleEvol still achieves the best performance score under 8 out of 10 backbone models on Taillard, indicating strong OOD competitiveness despite the negative model-intelligence trend.

\paragraph{Implication.}
Together, these results suggest that the effect of model intelligence depends on the evaluation distribution. On the synthetic IDD instances, stronger models can discover better heuristics with lower training-distribution gaps. However, the same improvements do not necessarily transfer to the OOD Taillard instances, whose structure differs from the synthetic search distribution in terms of job sizes and machine counts. In this sense, FSSP-GLS exposes a stronger benchmark-generalization gap than TSP and CVRP, where held-out test instances mainly vary in size while following the same uniform synthetic generation. We therefore treat the Taillard performance as a generalization test for evolved heuristics, rather than as a direct test of the AHI--ICE relationship. Recent work \cite{shi2026generalizable} attempts to address this issue by incorporating cross-size or cross-distributional instances into the training dataset, offering a complementary direction for improving generalization.


\subsection{Cost Analysis}
\label{sec:cost_analysis}

\begin{figure}[t]
\centering
{\small \textbf{(a) TSP Constructive}}

\vspace{0.2em}
\includegraphics[width=0.80\linewidth]{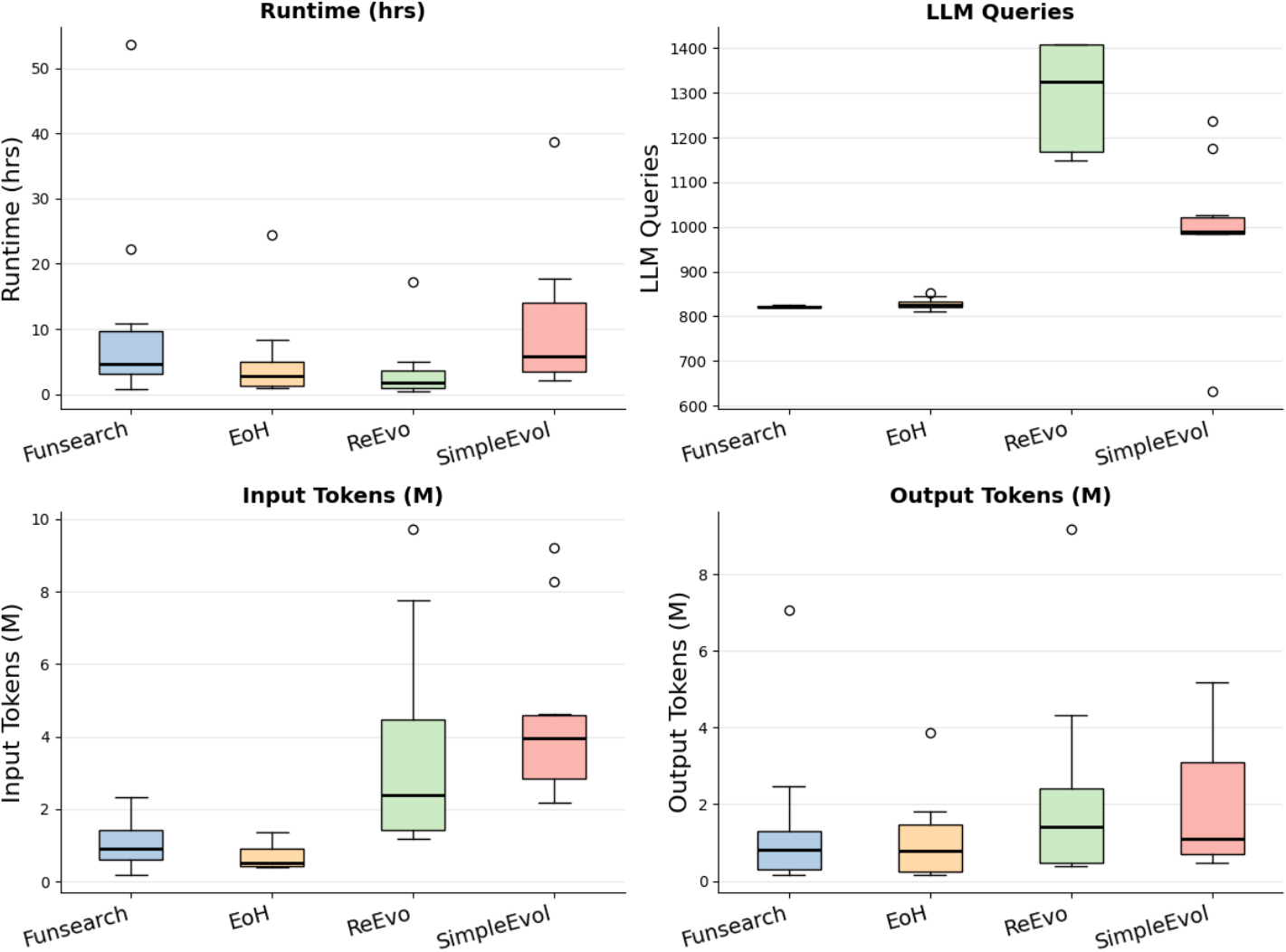}

\vspace{0.7em}

{\small \textbf{(b) CVRP-ACO}}

\vspace{0.2em}
\includegraphics[width=0.80\linewidth]{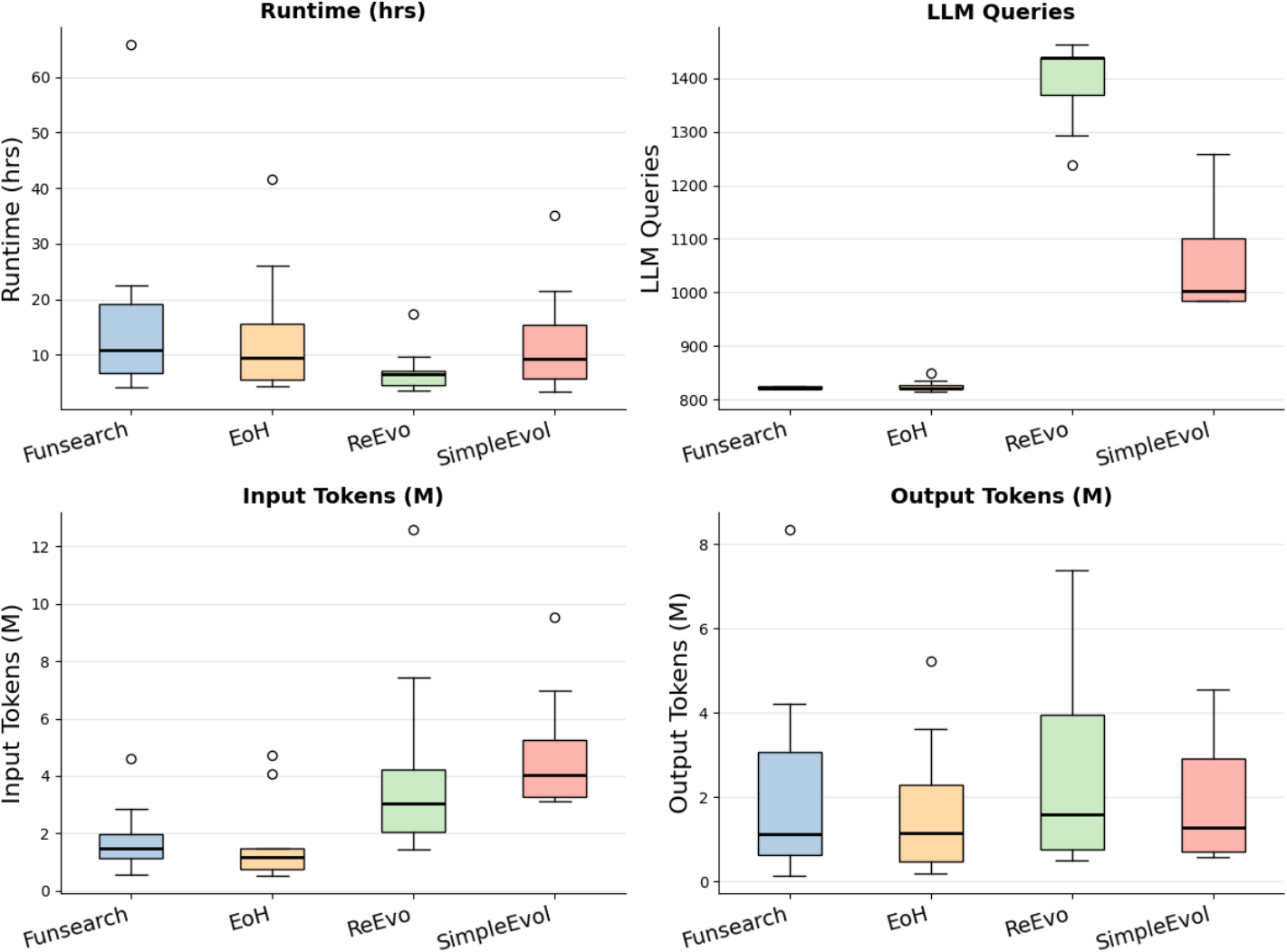}

\caption{Computational cost comparison of LLM-based AHD methods.}
\label{fig:cost_analysis}
\end{figure}

We report the computational cost of each AHD framework using training runtime, token consumption, and total LLM queries. All statistics are averaged over ten backbone models and three independent runs per model. The results are summarized in Table~\ref{tab:cost_analysis} and visualized in Figure~\ref{fig:cost_analysis}.

\paragraph{LLM queries.} ReEvo issues substantially more LLM queries than other frameworks, reflecting the design that requires separate calls on reflection generation, especially in crossover. FunSearch and EoH maintain query counts close to the number of evaluations (820--829), showing that their operators require fewer intermediate steps. SimpleEvol falls between these two groups, incurring a moderate number of additional calls primarily due to periodic history summarization.

\begin{table}[b]
\centering
\caption{Training time, token usage, and LLM query counts of all methods averaged across 10 models.}
\label{tab:cost_analysis}
\small
\setlength{\tabcolsep}{4pt}
\renewcommand{\arraystretch}{1.12}
\begin{tabular}{l|l|c|c|c|c}
\thickhline
\hline
Task & Method & Runtime (hrs) & Input Tokens (M) & Output Tokens (M) & \# Queries \\
\hline
\multirow{4}{*}{TSP Constructive}
& FunSearch & 11.13 & 1.111 & 1.479 & 821 \\
& EoH & 5.23 & 0.687 & 1.100 & 828 \\
& ReEvo & 3.60 & 3.602 & 2.249 & 1293 \\
& SimpleEvol (ours) & 10.55 & 4.517 & 1.988 & 1001 \\
\hline
\multirow{4}{*}{CVRP-ACO}
& FunSearch & 16.86 & 1.826 & 2.008 & 822 \\
& EoH & 13.86 & 1.693 & 1.570 & 825 \\
& ReEvo & 7.17 & 4.192 & 2.548 & 1397 \\
& SimpleEvol (ours) & 12.49 & 4.749 & 1.983 & 1055 \\
\hline
\thickhline
\end{tabular}
\end{table}
\paragraph{Token consumption.} SimpleEvol consumes the most input tokens among all frameworks, since its single-trajectory design retains a running context of meta-information and compressed summaries across iterations. However, its output token consumption remains moderate, given that the mean is lower than ReEvo on both tasks. This suggests that the additional input context does not translate into proportionally longer outputs. Since input tokens are priced substantially lower than output tokens in modern LLM APIs, the higher input consumption of SimpleEvol does not translate to a disproportionately higher monetary cost.

\paragraph{Runtime Consideration.} Training runtime varies considerably across frameworks and models. SimpleEvol's runtime is moderate on CVRP but somewhat higher on TSP Constructive. This is partly attributable to its single-trajectory nature. Unlike population-based methods such as EoH and ReEvo, SimpleEvol does not parallelize candidate generation and evaluation across multiple individuals, leading to longer sequential execution times on certain backbone models. The wide runtime distribution of all methods in Figure~\ref{fig:cost_analysis} also reflects the large variance in inference speed across backbone LLMs, particularly for the reasoning family.

\paragraph{Overall.} Taken together, these results indicate that SimpleEvol does not achieve its intelligence conversion advantage by consuming substantially more computational resources. Its query count and output token usage are comparable to or lower than ReEvo, which exhibits the lowest ICE among all evaluated frameworks. The primary cost overhead of SimpleEvol lies in input token consumption, which is a deliberate consequence of its context-rich single-trajectory design and carries relatively moderate monetary cost under standard API pricing.


\section{Detailed Performance Results for ICE Computation}
\label{sec:ice_raw_results}

To ensure transparency and reproducibility of the ICE computation, we report the detailed performance results for each AHD framework across all evaluated models.

Specifically, we provide the raw objective values and the corresponding optimality gaps in each test setting, along with the average gap across sizes and the derived performance score $P(m)$. These results serve as the basis for the ICE analysis presented in the main text. Note that the last three LLM models are reasoning models. The best results among all models for each problem size are marked in bold. All results are based on the average of three independent runs.

\begin{table*}[h]
\captionsetup{labelfont=bf}
\centering
\caption{Detailed performance of \textbf{SimpleEvol} across backbone models on the \textbf{TSP} test sets.}
\label{tab:simpleevol_detailed_results}
\setlength{\tabcolsep}{2.5pt}
\renewcommand{\arraystretch}{1.1}
\begin{tabular}{lcccccccc}
\thickhline
\multirow{2}{*}{LLM Model} 
& \multicolumn{2}{c}{\underline{$N=50$}}
& \multicolumn{2}{c}{$N=100$} 
& \multicolumn{2}{c}{$N=200$} 
& \multirow{2}{*}{Avg Gap$\downarrow$} 
& \multirow{2}{*}{$P(m)\uparrow$} \\
\cmidrule(lr){2-3} \cmidrule(lr){4-5} \cmidrule(lr){6-7}
& Obj. & Gap & Obj. & Gap & Obj. & Gap &  &  \\
\midrule
Opt (best known) & 5.6750 & -- & 7.7680 & -- & 10.6590 & -- & -- & 1.0000 \\
\midrule
GPT-4o-mini & 6.3438 & 11.79\% & 8.8025 & 13.32\% & 12.4151 & 16.48\% & 13.86\% & 7.2152 \\
GPT-4.1-nano & 6.2518 & 10.16\% & 8.7732 & 12.94\% & 12.2825 & 15.23\% & 12.78\% & 7.8256 \\
Claude Sonnet 3.7 & 6.1693 & 8.71\% & 8.7684 & 12.88\% & 12.4542 & 16.84\% & 12.81\% & 7.8065 \\
GPT-4.1-mini & 6.3079 & 11.15\% & 8.8527 & 13.96\% & 12.3860 & 16.20\% & 13.77\% & 7.2609 \\
DeepSeek-v3 & 6.3868 & 12.54\% & 8.9293 & 14.95\% & 12.5136 & 17.40\% & 14.96\% & 6.6826 \\
Gemini-2.5-Flash & 6.2126 & 9.47\% & 8.8990 & 14.56\% & 12.5300 & 17.55\% & 13.86\% & 7.2139 \\
Qwen3-235B-Instruct & 6.3036 & 11.08\% & 8.7748 & 12.96\% & 12.3325 & 15.70\% & 13.25\% & 7.5496 \\
\midrule
o3-mini & 6.2156 & 9.53\% & 8.6338 & 11.15\% & 12.0638 & 13.18\% & 11.28\% & 8.8625 \\
Qwen3-235B-Thinking & 6.1917 & 9.10\% & 8.7295 & 12.38\% & 12.3152 & 15.54\% & 12.34\% & 8.1035\\
GPT-5-mini & \textbf{5.9455} & \textbf{4.77\%} & \textbf{8.2706} & \textbf{6.47\%} & \textbf{11.6710} & \textbf{9.49\%} & \textbf{6.91\%} & \textbf{14.4706} \\
\midrule
AVG & 6.2329 & 9.83\% & 8.7434 & 12.56\% & 12.2964 & 15.36\% & 12.58\% & -- \\
\thickhline
\end{tabular}
\end{table*}

\begin{table*}[h]
\captionsetup{labelfont=bf}
\centering
\caption{Detailed performance of \textbf{FunSearch} across backbone models on the \textbf{TSP} test sets.}
\label{tab:funsearch_detailed_results}
\setlength{\tabcolsep}{2.5pt}
\renewcommand{\arraystretch}{1.1}
\begin{tabular}{lcccccccc}
\thickhline
\multirow{2}{*}{LLM Model} 
& \multicolumn{2}{c}{\underline{$N=50$}}
& \multicolumn{2}{c}{$N=100$} 
& \multicolumn{2}{c}{$N=200$} 
& \multirow{2}{*}{Avg Gap$\downarrow$} 
& \multirow{2}{*}{$P(m)\uparrow$} \\
\cmidrule(lr){2-3} \cmidrule(lr){4-5} \cmidrule(lr){6-7}
& Obj. & Gap & Obj. & Gap & Obj. & Gap &  &  \\
\midrule
Opt (best known) & 5.6750 & -- & 7.7680 & -- & 10.6590 & -- & -- & 1.0000 \\
\midrule
GPT-4o-mini & 6.3942 & 12.67\% & 8.8689 & 14.17\% & 12.4607 & 16.90\% & 14.58\% & 6.8574 \\
GPT-4.1-nano & 6.3294 & 11.53\% & 8.8852 & 14.38\% & 12.3365 & 15.74\% & 13.88\% & 7.2029 \\
Claude Sonnet 3.7 & 6.1857 & 9.00\% & 8.7873 & 13.12\% & 12.4140 & 16.47\% & 12.86\% & 7.7749 \\
GPT-4.1-mini & 6.3335 & 11.60\% & 8.8362 & 13.75\% & 12.3895 & 16.24\% & 13.86\% & 7.2132 \\
DeepSeek-v3 & 6.3620 & 12.11\% & 8.7398 & 12.51\% & 12.2927 & 15.33\% & 13.31\% & 7.5105 \\
Gemini-2.5-Flash & 6.3410 & 11.74\% & 8.7475 & 12.61\% & 12.2027 & 14.48\% & 12.94\% & 7.7265 \\
Qwen3-235B-Instruct & 6.3066 & 11.13\% & 8.7764 & 12.98\% & 12.4095 & 16.42\% & 13.51\% & 7.4013 \\
\midrule
o3-mini & 6.2896 & 10.83\% & 8.7349 & 12.45\% & 12.2437 & 14.87\% & 12.71\% & 7.8650 \\
Qwen3-235B-Thinking & 6.2562 & 10.24\% & 8.7058 & 12.07\% & 12.1561 & 14.05\% & 12.12\% & 8.2508 \\
GPT-5-mini & \textbf{5.9869} & \textbf{5.50\%} & \textbf{8.3378} & \textbf{7.33\%} & \textbf{11.7617} & \textbf{10.35\%} & \textbf{7.73\%} & \textbf{12.9449} \\
\midrule
AVG & 6.2785 & 10.63\% & 8.7420 & 12.54\% & 12.2667 & 15.08\% & 12.75\% & -- \\
\thickhline
\end{tabular}
\end{table*}

\begin{table*}[h]
\captionsetup{labelfont=bf}
\centering
\caption{Detailed performance of \textbf{EoH} across backbone models on the \textbf{TSP} test sets.}
\label{tab:eoh_detailed_results}
\setlength{\tabcolsep}{2.5pt}
\renewcommand{\arraystretch}{1.1}
\begin{tabular}{lcccccccc}
\thickhline

\multirow{2}{*}{LLM Model} 
& \multicolumn{2}{c}{\underline{$N=50$}} 
& \multicolumn{2}{c}{$N=100$} 
& \multicolumn{2}{c}{$N=200$} 
& \multirow{2}{*}{Avg Gap$\downarrow$} 
& \multirow{2}{*}{$P(m)\uparrow$} \\
\cmidrule(lr){2-3} \cmidrule(lr){4-5} \cmidrule(lr){6-7}
& Obj. & Gap & Obj. & Gap & Obj. & Gap &   &  \\
\midrule
Opt (best known) & 5.6750 & -- & 7.7680 & -- & 10.6590 & --  & -- & 1.0000 \\
\midrule
GPT-4o-mini & 6.5120 & 14.75\% & 9.1810 & 18.19\% & 12.7930 & 20.02\% & 17.65\% & 5.6647 \\
GPT-4.1-nano & 6.2992 & 11.00\% & 8.7966 & 13.24\% & 12.2683 & 15.10\% & 13.11\% & 7.6261 \\
Claude Sonnet 3.7 & 6.2576 & 10.27\% & 8.7952 & 13.22\% & 12.5031 & 17.30\% & 13.60\% & 7.3546 \\
GPT-4.1-mini & 6.3993 & 12.76\% & 8.9158 & 14.78\% & 12.6525 & 18.70\% & 15.41\% & 6.4877 \\
DeepSeek-v3 & 6.4688 & 13.99\% & 9.0123 & 16.02\% & 12.6913 & 19.07\% & 16.36\% & 6.1134 \\
Gemini-2.5-Flash & 6.2702 & 10.49\% & 8.7831 & 13.07\% & 12.2965 & 15.36\% & 12.97\% & 7.7084 \\
Qwen3-235B-Instruct & 6.5616 & 15.62\% & 9.0655 & 16.70\% & 12.7272 & 19.40\% & 17.24\% & 5.7995 \\
\midrule
o3-mini & 6.2471 & 10.08\% & 8.8024 & 13.32\% & 12.3182 & 15.57\% & 12.99\% & 7.6995 \\
Qwen3-235B-Thinking & 6.2754 & 10.58\% & 8.8831 & 14.36\% & 12.5551 & 17.79\% & 14.24\% & 7.0219 \\
GPT-5-mini & \textbf{6.0588} & \textbf{6.76\%} & \textbf{8.4234} & \textbf{8.44\%} & \textbf{11.7847} & \textbf{10.56\%} & \textbf{8.59\%} & \textbf{11.6451} \\
\midrule
AVG & 6.3350 & 11.63\% & 8.8658 & 14.13\% & 12.4590 & 16.89\% & 14.22\% & -- \\
\thickhline
\end{tabular}
\end{table*}

\begin{table*}[h]
\captionsetup{labelfont=bf}
\centering
\caption{Detailed performance of \textbf{ReEvo} across backbone models on the \textbf{TSP} test sets.}
\label{tab:reevo_detailed_results}
\setlength{\tabcolsep}{2.5pt}
\renewcommand{\arraystretch}{1.1}
\begin{tabular}{lcccccccc}
\thickhline
\multirow{2}{*}{LLM Model} 
& \multicolumn{2}{c}{\underline{$N=50$} }
& \multicolumn{2}{c}{$N=100$} 
& \multicolumn{2}{c}{$N=200$} 
& \multirow{2}{*}{Avg Gap$\downarrow$} 
& \multirow{2}{*}{$P(m)\uparrow$} \\
\cmidrule(lr){2-3} \cmidrule(lr){4-5} \cmidrule(lr){6-7}
& Obj. & Gap & Obj. & Gap & Obj. & Gap &  &  \\
\midrule
Opt (best known) & 5.6750 & -- & 7.7680 & -- & 10.6590 & -- & -- & 1.0000 \\
\midrule
GPT-4o-mini & 6.3526 & 11.94\% & 8.8555 & 14.00\% & 12.4488 & 16.79\% & 14.24\% & 7.0207 \\
GPT-4.1-nano & 6.3886 & 12.58\% & 9.0631 & 16.67\% & 12.6207 & 18.40\% & 15.88\% & 6.2957 \\
Claude Sonnet 3.7 & 6.2048 & 9.34\% & 8.6615 & 11.50\% & 12.2265 & 14.71\% & 11.85\% & 8.4402 \\
GPT-4.1-mini & 6.5783 & 15.92\% & 9.2570 & 19.17\% & 13.2415 & 24.23\% & 19.77\% & 5.0580 \\
DeepSeek-v3 & 6.4249 & 13.21\% & 9.0256 & 16.19\% & 12.6479 & 18.66\% & 16.02\% & 6.2418 \\
Gemini-2.5-Flash & 6.1797 & 8.89\% & 8.6949 & 11.93\% & 12.2727 & 15.14\% & 11.99\% & 8.3416 \\
Qwen3-235B-Instruct & 6.5162 & 14.82\% & 9.0987 & 17.13\% & 12.7149 & 19.29\% & 17.08\% & 5.8548 \\
\midrule
o3-mini  & 6.3930 & 12.65\% & 9.0021 & 15.89\% & 12.5043 & 17.31\% & 15.28\% & 6.5430 \\
Qwen3-235B-Thinking & 6.2105 & 9.44\% & 8.7497 & 12.64\% & 12.3568 & 15.93\% & 12.67\% & 7.8941 \\
GPT-5-mini & \textbf{6.1280} & \textbf{7.98\%} & \textbf{8.5494} & \textbf{10.06\%} & \textbf{11.9487} & \textbf{12.10\%} & \textbf{10.05\%} & \textbf{9.9533} \\
\midrule
AVG & 6.3376 & 11.68\% & 8.8957 & 14.52\% & 12.4983 & 17.26\% & 14.48\% & -- \\
\thickhline
\end{tabular}
\end{table*}

\begin{table*}[h]
\captionsetup{labelfont=bf}
\centering
\caption{Detailed performance of \textbf{MCTS-AHD} across backbone models on the \textbf{TSP} test sets.}
\label{tab:mcts_detailed_results}
\setlength{\tabcolsep}{2.5pt}
\renewcommand{\arraystretch}{1.1}
\begin{tabular}{lcccccccc}
\thickhline
\multirow{2}{*}{LLM Model} 
& \multicolumn{2}{c}{\underline{$N=50$}}
& \multicolumn{2}{c}{$N=100$} 
& \multicolumn{2}{c}{$N=200$} 
& \multirow{2}{*}{Avg Gap$\downarrow$} 
& \multirow{2}{*}{$P(m)\uparrow$} \\
\cmidrule(lr){2-3} \cmidrule(lr){4-5} \cmidrule(lr){6-7}
& Obj. & Gap & Obj. & Gap & Obj. & Gap &  &  \\
\midrule
Opt (best known) & 5.6750 & -- & 7.7680 & -- & 10.6590 & -- & -- & 1.0000 \\
\midrule
GPT-4o-mini 
& 6.2326 & 9.83\% 
& 8.7412 & 12.53\% 
& 12.2513 & 14.94\% 
& 12.43\% & 8.0446 \\

GPT-4.1-nano 
& 6.2886 & 10.81\% 
& 9.0136 & 16.04\% 
& 12.5707 & 17.94\% 
& 14.93\% & 6.6989 \\

Claude Sonnet 3.7 
& 6.2341 & 9.85\% 
& 8.6815 & 11.76\% 
& 12.2565 & 14.99\% 
& 12.20\% & 8.1968 \\

GPT-4.1-mini 
& 6.3610 & 12.09\% 
& 9.1713 & 18.07\% 
& 13.0820 & 22.73\% 
& 17.63\% & 5.6727 \\

DeepSeek-v3 
& 6.3949 & 12.68\% 
& 8.9548 & 15.28\% 
& 12.4796 & 17.08\% 
& 15.01\% & 6.6602 \\

Gemini-2.5-Flash 
& 6.2813 & 10.68\% 
& 8.7210 & 12.27\% 
& 12.3124 & 15.51\% 
& 12.82\% & 7.7996 \\

Qwen3-235B-Instruct 
& 6.3713 & 12.27\% 
& 8.8913 & 14.46\% 
& 12.4857 & 17.14\% 
& 14.62\% & 6.8387 \\

\midrule
o3-mini 
& 6.3230 & 11.42\% 
& 8.7951 & 13.22\% 
& 12.3629 & 15.99\% 
& 13.54\% & 7.3844 \\

Qwen3-235B-Thinking 
& 6.2150 & 9.52\% 
& 8.6974 & 11.96\% 
& 12.4024 & 16.36\% 
& 12.61\% & 7.9290 \\

GPT-5-mini 
& \textbf{6.1212} & \textbf{7.86\%} 
& \textbf{8.5312} & \textbf{9.83\%} 
& \textbf{11.8913} & \textbf{11.56\%} 
& \textbf{9.75\%} & \textbf{10.2568} \\

\midrule
AVG 
& 6.2823 & 10.70\% 
& 8.8199 & 13.54\% 
& 12.4095 & 16.42\% 
& 13.55\% & -- \\
\thickhline
\end{tabular}
\end{table*}

\newpage

\begin{table*}[h]
\captionsetup{labelfont=bf}
\centering
\caption{Detailed performance of \textbf{SimpleEvol} across backbone models on the \textbf{CVRP} test sets.}
\label{tab:simpleevol_cvrp_detailed_results}
\setlength{\tabcolsep}{2.5pt}
\renewcommand{\arraystretch}{1.1}
\begin{tabular}{lcccccccc}
\thickhline
\multirow{2}{*}{LLM Model} 
& \multicolumn{2}{c}{\underline{$N=50$} }
& \multicolumn{2}{c}{$N=100$} 
& \multicolumn{2}{c}{$N=200$} 
& \multirow{2}{*}{Avg Gap$\downarrow$} 
& \multirow{2}{*}{$P(m)\uparrow$} \\
\cmidrule(lr){2-3} \cmidrule(lr){4-5} \cmidrule(lr){6-7}
& Obj. & Gap & Obj. & Gap & Obj. & Gap &  &  \\
\midrule
Opt (best known) & 8.8880 & -- & 14.9320 & -- & 27.1590 & -- & -- & 1.0000 \\
\midrule
GPT-4o-mini & 9.2702 & 4.30\% & 16.1382 & 8.08\% & 28.6203 & 5.38\% & 5.92\% & 16.8932 \\
GPT-4.1-nano & 9.1717 & 3.19\% & 16.1514 & 8.17\% & 28.9365 & 6.54\% & 5.97\% & 16.7569 \\
Claude Sonnet 3.7 & 9.2351 & 3.90\% & 15.9603 & 6.89\% & 28.3293 & 4.31\% & 5.03\% & 19.8669 \\
GPT-4.1-mini & 9.1025 & 2.41\% & 16.1855 & 8.39\% & 29.1563 & 7.35\% & 6.05\% & 16.5178 \\
DeepSeek-v3 & 9.0888 & 2.26\% & 16.1162 & 7.93\% & 28.8299 & 6.15\% & 5.45\% & 18.3575 \\
Gemini-2.5-Flash & 9.0263 & 1.56\% & 16.1653 & 8.26\% & 29.2523 & 7.71\% & 5.84\% & 17.1203 \\
Qwen3-235B-Instruct & 8.9668 & 0.89\% & 15.7612 & 5.55\% & 28.3393 & 4.35\% & 3.60\% & 27.8147 \\
\midrule
o3-mini & 9.2048 & 3.56\% & 16.1552 & 8.19\% & 28.6612 & 5.53\% & 5.76\% & 17.3538 \\
Qwen3-235B-Thinking & 8.9433 & 0.62\% & 15.9917 & 7.10\% & 28.4568 & 4.78\% & 4.17\% & 24.0047 \\
GPT-5-mini & \textbf{8.9183} & \textbf{0.34\%} & \textbf{15.5761} & \textbf{4.31\%} & \textbf{28.3167} & \textbf{4.26\%} & \textbf{2.97\%} & \textbf{33.6431} \\
\midrule
AVG & 9.0928 & 2.30\% & 16.0201 & 7.29\% & 28.6899 & 5.64\% & 5.08\% & -- \\
\thickhline
\end{tabular}
\end{table*}

\begin{table*}[h]
\captionsetup{labelfont=bf}
\centering
\caption{Detailed performance of \textbf{FunSearch} across backbone models on the \textbf{CVRP} test sets.}
\label{tab:funsearch_cvrp_detailed_results}
\setlength{\tabcolsep}{2.5pt}
\renewcommand{\arraystretch}{1.1}
\begin{tabular}{lcccccccc}
\thickhline
\multirow{2}{*}{LLM Model} 
& \multicolumn{2}{c}{\underline{$N=50$} }
& \multicolumn{2}{c}{$N=100$} 
& \multicolumn{2}{c}{$N=200$} 
& \multirow{2}{*}{Avg Gap$\downarrow$} 
& \multirow{2}{*}{$P(m)\uparrow$} \\
\cmidrule(lr){2-3} \cmidrule(lr){4-5} \cmidrule(lr){6-7}
& Obj. & Gap & Obj. & Gap & Obj. & Gap &  &  \\
\midrule
Opt (best known) & 8.8880 & -- & 14.9320 & -- & 27.1590 & -- & -- & 1.0000 \\
\midrule
GPT-4o-mini & 9.2106 & 3.63\% & 16.5272 & 10.68\% & 29.5915 & 8.96\% & 7.76\% & 12.8926 \\
GPT-4.1-nano & 9.3651 & 5.37\% & 16.5791 & 11.03\% & 29.1303 & 7.26\% & 7.89\% & 12.6813 \\
Claude Sonnet 3.7 & 9.0786 & 2.14\% & 16.1344 & 8.05\% & 28.8580 & 6.26\% & 5.48\% & 18.2346 \\
GPT-4.1-mini & 9.1347 & 2.78\% & 16.6122 & 11.25\% & 29.7963 & 9.71\% & 7.91\% & 12.6377 \\
DeepSeek-v3 & \textbf{8.9294} & \textbf{0.47\%} & 15.9658 & 6.92\% & 28.5458 & 5.11\% & 4.17\% & 24.0088 \\
Gemini-2.5-Flash & 9.2618 & 4.21\% & 16.4287 & 10.02\% & 29.5897 & 8.95\% & 7.73\% & 12.9428 \\
Qwen3-235B-Instruct & 9.0032 & 1.30\% & 16.1471 & 8.14\% & 28.7328 & 5.79\% & 5.08\% & 19.6993 \\
\midrule
o3-mini & 9.0320 & 1.62\% & 16.2841 & 9.06\% & 29.1465 & 7.32\% & 6.00\% & 16.6729 \\
Qwen3-235B-Thinking & 9.0272 & 1.57\% & 16.1238 & 7.98\% & 28.6510 & 5.49\% & 5.01\% & 19.9452 \\
GPT-5-mini & 8.9802 & 1.04\% & \textbf{15.6525} & \textbf{4.83\%} & \textbf{28.3694} & \textbf{4.46\%} & \textbf{3.44\%} & \textbf{29.0718} \\
\midrule
AVG & 9.1023 & 2.41\% & 16.2455 & 8.80\% & 29.0411 & 6.93\% & 6.05\% & -- \\
\thickhline
\end{tabular}
\end{table*}

\begin{table*}[h]
\captionsetup{labelfont=bf}
\centering
\caption{Detailed performance of \textbf{EoH} across backbone models on the \textbf{CVRP} test sets.}
\label{tab:eoh_cvrp_detailed_results}
\setlength{\tabcolsep}{2.5pt}
\renewcommand{\arraystretch}{1.1}
\begin{tabular}{lcccccccc}
\thickhline
\multirow{2}{*}{LLM Model} 
& \multicolumn{2}{c}{\underline{$N=50$} }
& \multicolumn{2}{c}{$N=100$} 
& \multicolumn{2}{c}{$N=200$} 
& \multirow{2}{*}{Avg Gap$\downarrow$} 
& \multirow{2}{*}{$P(m)\uparrow$} \\
\cmidrule(lr){2-3} \cmidrule(lr){4-5} \cmidrule(lr){6-7}
& Obj. & Gap & Obj. & Gap & Obj. & Gap &  &  \\
\midrule
Opt (best known) & 8.8880 & -- & 14.9320 & -- & 27.1590 & -- & -- & 1.0000 \\
\midrule
GPT-4o-mini & 9.1090 & 2.49\% & 16.1046 & 7.85\% & 28.4425 & 4.73\% & 5.02\% & 19.9131 \\
GPT-4.1-nano & 9.2774 & 4.38\% & 16.5862 & 11.08\% & 29.2250 & 7.61\% & 7.69\% & 13.0055 \\
Claude Sonnet 3.7 & 8.9577 & 0.78\% & 15.8456 & 6.12\% & 28.6107 & 5.35\% & 4.08\% & 24.4939 \\
GPT-4.1-mini & 9.5867 & 7.86\% & 16.3944 & 9.79\% & 28.8408 & 6.19\% & 7.95\% & 12.5804 \\
DeepSeek-v3 & 9.1249 & 2.67\% & 16.3852 & 9.73\% & 29.2224 & 7.60\% & 6.67\% & 15.0031 \\
Gemini-2.5-Flash & 9.1466 & 2.91\% & 16.3306 & 9.37\% & 29.3662 & 8.13\% & 6.80\% & 14.7038 \\
Qwen3-235B-Instruct & 8.9705 & 0.93\% & \textbf{15.8090} & \textbf{5.87\%} & 28.4138 & 4.62\% & \textbf{3.81\%} & \textbf{26.2668} \\
\midrule
o3-mini & 9.2051 & 3.57\% & 15.8788 & 6.34\% & 28.5589 & 5.15\% & 5.02\% & 19.9164 \\
Qwen3-235B-Thinking & \textbf{8.9239} & \textbf{0.40\%} & 16.0438 & 7.45\% & 28.6985 & 5.67\% & 4.51\% & 22.1931 \\
GPT-5-mini & 8.9507 & 0.71\% & 15.9324 & 6.70\% & \textbf{28.3221} & \textbf{4.28\%} & 3.90\% & 25.6674 \\
\midrule
AVG & 9.1253 & 2.67\% & 16.1311 & 8.03\% & 28.7701 & 5.93\% & 5.54\% & -- \\
\thickhline
\end{tabular}
\end{table*}

\begin{table*}[h]
\captionsetup{labelfont=bf}
\centering
\caption{Detailed performance of \textbf{ReEvo} across backbone models on the \textbf{CVRP} test sets.}
\label{tab:reevo_cvrp_detailed_results}
\setlength{\tabcolsep}{2.5pt}
\renewcommand{\arraystretch}{1.1}
\begin{tabular}{lcccccccc}
\thickhline
\multirow{2}{*}{LLM Model} 
& \multicolumn{2}{c}{\underline{$N=50$} }
& \multicolumn{2}{c}{$N=100$} 
& \multicolumn{2}{c}{$N=200$} 
& \multirow{2}{*}{Avg Gap$\downarrow$} 
& \multirow{2}{*}{$P(m)\uparrow$} \\
\cmidrule(lr){2-3} \cmidrule(lr){4-5} \cmidrule(lr){6-7}
& Obj. & Gap & Obj. & Gap & Obj. & Gap &  &  \\
\midrule
Opt (best known) & 8.8880 & -- & 14.9320 & -- & 27.1590 & -- & -- & 1.0000 \\
\midrule
GPT-4o-mini & 9.3572 & 5.28\% & 16.1092 & 7.88\% & 29.2078 & 7.54\% & 6.90\% & 14.4882 \\
GPT-4.1-nano & 9.1683 & 3.15\% & 16.1404 & 8.09\% & 28.9847 & 6.72\% & 5.99\% & 16.6958 \\
Claude Sonnet 3.7 & 9.0870 & 2.24\% & 15.9228 & 6.64\% & \textbf{28.2793} & \textbf{4.12\%} & 4.33\% & 23.0788 \\
GPT-4.1-mini & 9.1950 & 3.45\% & 16.0215 & 7.30\% & 28.7532 & 5.87\% & 5.54\% & 18.0501 \\
DeepSeek-v3 & 8.9550 & 0.75\% & 15.9563 & 6.86\% & 28.5223 & 5.02\% & 4.21\% & 23.7468 \\
Gemini-2.5-Flash & 9.0277 & 1.57\% & 16.2299 & 8.69\% & 29.1971 & 7.50\% & 5.92\% & 16.8837 \\
Qwen3-235B-Instruct & 9.2608 & 4.19\% & 16.2857 & 9.07\% & 28.9658 & 6.65\% & 6.64\% & 15.0658 \\
\midrule
o3-mini & 9.2917 & 4.54\% & 16.5532 & 10.86\% & 29.5430 & 8.78\% & 8.06\% & 12.4081 \\
Qwen3-235B-Thinking & 8.9858 & 1.10\% & 16.0516 & 7.50\% & 28.6316 & 5.42\% & 4.67\% & 21.3972 \\
GPT-5-mini & \textbf{8.9315} & \textbf{0.49\%} & \textbf{15.8244} & \textbf{5.98\%} & 28.3480 & 4.38\% & \textbf{3.61\%} & \textbf{27.6656} \\
\midrule
AVG & 9.1260 & 2.68\% & 16.1095 & 7.89\% & 28.8433 & 6.20\% & 5.59\% & -- \\
\thickhline
\end{tabular}
\end{table*}

\begin{table*}[h]
\captionsetup{labelfont=bf}
\centering
\caption{Detailed performance of \textbf{MCTS-AHD} across backbone models on the \textbf{CVRP} test sets.}
\label{tab:mcts_cvrp_detailed_results}
\setlength{\tabcolsep}{2.5pt}
\renewcommand{\arraystretch}{1.1}
\begin{tabular}{lcccccccc}
\thickhline
\multirow{2}{*}{LLM Model} 
& \multicolumn{2}{c}{\underline{$N=50$}}
& \multicolumn{2}{c}{$N=100$} 
& \multicolumn{2}{c}{$N=200$} 
& \multirow{2}{*}{Avg Gap$\downarrow$} 
& \multirow{2}{*}{$P(m)\uparrow$} \\
\cmidrule(lr){2-3} \cmidrule(lr){4-5} \cmidrule(lr){6-7}
& Obj. & Gap & Obj. & Gap & Obj. & Gap &  &  \\
\midrule
Opt (best known) & 8.8880 & -- & 14.9320 & -- & 27.1590 & -- & -- & 1.0000 \\
\midrule
GPT-4o-mini 
& 9.2480 & 4.05\% 
& 15.7320 & 5.36\% 
& 28.3610 & 4.43\% 
& 4.61\% & 21.6855 \\

GPT-4.1-nano 
& 9.1572 & 3.03\% 
& 16.0282 & 7.34\% 
& 29.1074 & 7.17\% 
& 5.85\% & 17.0997 \\

Claude Sonnet 3.7 
& 9.1691 & 3.16\% 
& 15.7503 & 5.48\% 
& \textbf{28.2169} & \textbf{3.90\%} 
& 4.18\% & 23.9271 \\

GPT-4.1-mini 
& 9.0253 & 1.54\% 
& 15.8324 & 6.03\% 
& 28.5718 & 5.20\% 
& 4.26\% & 23.4802 \\

DeepSeek-v3 
& 9.1002 & 2.39\% 
& 15.8325 & 6.03\% 
& 28.6238 & 5.39\% 
& 4.60\% & 21.7209 \\

Gemini-2.5-Flash 
& 9.2131 & 3.66\% 
& 16.3104 & 9.23\% 
& 29.4641 & 8.49\% 
& 7.13\% & 14.0342 \\

Qwen3-235B-Instruct 
& 9.0737 & 2.09\% 
& \textbf{15.6812} & \textbf{5.02\%} 
& 28.3809 & 4.50\% 
& 3.87\% & 25.8491 \\

\midrule
o3-mini 
& 9.1982 & 3.49\% 
& 16.1208 & 7.96\% 
& 28.8421 & 6.20\% 
& 5.88\% & 16.9984 \\

Qwen3-235B-Thinking & \textbf{8.9821} & \textbf{1.06\%} & 15.7857 & 5.72\% & 28.7939 & 6.02\% & 4.27\% & 23.4453 \\

GPT-5-mini 
& 8.9849 & 1.09\%
& 15.7011 & 5.15\% 
& 28.4893 & 4.90\% 
& \textbf{3.71\%} & \textbf{26.9321} \\

\midrule
AVG & 9.1152 & 2.56\% & 15.8775 & 6.33\% & 28.6851 & 5.62\% & 4.84\% & -- \\
\thickhline
\end{tabular}
\end{table*}

\newpage

\begin{table*}[h]
\captionsetup{labelfont=bf}
\centering
\caption{Performance comparison across AHD frameworks on IDD and Taillard (OOD) instances.}
\label{tab:idd_ood_comparison_fssp}
\setlength{\tabcolsep}{6pt}
\renewcommand{\arraystretch}{1.03}
\begin{tabular}{lcccc}
\thickhline
\thickhline
LLM Model 
& Avg Gap (IDD)$\downarrow$ 
& $P(m)_{\text{idd}}\uparrow$ 
& Avg Gap (Taillard)$\downarrow$ 
& $P(m)_{\text{OOD}}\uparrow$ \\
\thickhline

\multicolumn{5}{c}{\textbf{LLM-based AHD: FunSearch}} \\
\midrule
GPT-4o-mini & 3.07\% & 32.5862 & 0.33\% & 307.5031 \\
GPT-4.1-nano & 3.02\% & 33.1680 & 0.27\% & 366.1662 \\
Claude Sonnet 3.7 & 3.00\% & 33.3845 & 0.28\% & 353.2321 \\
DeepSeek-v3 & 2.93\% & 34.0887 & 0.28\% & 351.9887 \\
GPT-4.1-mini & 2.93\% & 34.1319 & 0.41\% & 244.9180 \\
Gemini-2.5-Flash & 2.89\% & 34.5727 & 0.27\% & 375.2345 \\
Qwen3-235B-Instruct & 3.03\% & 32.9781 & 0.47\% & 215.0154 \\
o3-mini & 2.96\% & 33.7690 & 0.33\% & 306.4383 \\
Qwen3-235B-Thinking & 2.99\% & 33.4010 & 1.35\% & 74.1586 \\
GPT-5-mini & 2.94\% & 33.9986 & 0.35\% & 284.7380 \\
\midrule
\textbf{AVG} & 2.98\% & 33.60 & 0.43\% & 230.87 \\
\midrule

\multicolumn{5}{c}{\textbf{LLM-based AHD: EoH}} \\
\midrule
GPT-4o-mini & 3.15\% & 31.7396 & 0.37\% & 268.1720 \\
GPT-4.1-nano & 2.99\% & 33.4796 & 0.30\% & 338.5298 \\
Claude Sonnet 3.7 & 2.94\% & 33.9789 & 0.29\% & 347.7051 \\
DeepSeek-v3 & 2.95\% & 33.8593 & 0.27\% & 375.5925 \\
GPT-4.1-mini & 2.93\% & 34.0719 & 0.34\% & 296.9089 \\
Gemini-2.5-Flash & 2.96\% & 33.7680 & 0.29\% & 345.6787 \\
Qwen3-235B-Instruct & 3.08\% & 32.4926 & 0.48\% & 207.9780 \\
o3-mini & 2.95\% & 33.8673 & 0.33\% & 306.4383 \\
Qwen3-235B-Thinking & 3.03\% & 32.9837 & 1.37\% & 72.8527 \\
GPT-5-mini & 2.99\% & 33.4796 & 0.35\% & 284.7380 \\
\midrule
\textbf{AVG} & 3.00\% & 33.36 & 0.44\% & 228.35 \\
\midrule

\multicolumn{5}{c}{\textbf{LLM-based AHD: ReEvo}} \\
\midrule
GPT-4o-mini & 3.04\% & 32.9353 & 1.03\% & 96.8153 \\
GPT-4.1-nano & 2.91\% & 34.3833 & 1.08\% & 92.9891 \\
Claude Sonnet 3.7 & 2.98\% & 33.5289 & 0.90\% & 111.6944 \\
DeepSeek-v3 & 2.90\% & 34.5387 & 0.94\% & 106.8095 \\
GPT-4.1-mini & 2.86\% & 34.9876 & 1.07\% & 93.6298 \\
Gemini-2.5-Flash & 2.93\% & 34.1000 & 1.11\% & 90.1481 \\
Qwen3-235B-Instruct & 3.02\% & 33.0736 & 1.26\% & 79.3135 \\
o3-mini & 2.90\% & 34.5295 & 1.21\% & 82.8432 \\
Qwen3-235B-Thinking & 2.98\% & 33.5135 & 1.24\% & 80.8996 \\
GPT-5-mini & 2.94\% & 33.9766 & 0.91\% & 110.2657 \\
\midrule
\textbf{AVG} & \textbf{2.95\%} & \textbf{33.94} & 1.07\% & 93.21 \\
\midrule

\multicolumn{5}{c}{\textbf{LLM-based AHD: SimpleEvol}} \\
\midrule
GPT-4o-mini & 3.13\% & 31.9424 & 0.29\% & 340.8316 \\
GPT-4.1-nano & 2.94\% & 34.0348 & 0.27\% & 371.1952 \\
Claude Sonnet 3.7 & 2.99\% & 33.4717 & 0.28\% & 361.2717 \\
DeepSeek-v3 & 2.89\% & 34.5614 & 0.24\% & 413.9073 \\
GPT-4.1-mini & 2.93\% & 34.1767 & 0.31\% & 319.8465 \\
Gemini-2.5-Flash & 2.86\% & 34.9481 & 0.28\% & 352.2367 \\
Qwen3-235B-Instruct & 2.98\% & 33.5838 & 0.60\% & 167.6446 \\
o3-mini & 2.92\% & 34.2681 & 0.31\% & 325.6268 \\
Qwen3-235B-Thinking & 2.98\% & 33.5326 & 1.12\% & 89.0710 \\
GPT-5-mini & 2.94\% & 34.0646 & 0.31\% & 325.8390 \\
\midrule
\textbf{AVG} & 2.96\% & 33.84 & \textbf{0.40\%} & \textbf{249.32} \\
\thickhline
\thickhline
\end{tabular}
\end{table*}
\clearpage

\section{Limitations and Future Work}
\label{sec:limitations}
Our work establishes an empirical relationship between framework complexity and intelligence conversion efficiency. The design and ablation of SimpleEvol provide initial evidence that specific handcrafted structures such as constrained, rule-based operators can limit LLM autonomy, while self-directed designs such as LLM-based planning and summarization can more efficiently convert model intelligence into heuristic quality. However, it remains an open question whether the observed AHI–ICE relationship generalizes to other heuristic design domains, such as hyperparameter optimization or algorithm configuration, where evaluation feedback is stochastic or mediated by a surrogate rather than a deterministic solver. In future work, we plan to broaden the scope of evaluation to a larger set of domains beyond combinatorial optimization, so as to further validate the intelligence conversion perspective introduced here.

\section{Broader Impact}
\label{sec:impact_statement}
This work studies how the design of automated heuristic design (AHD) frameworks affects the efficiency of converting LLM capabilities into optimization performance, and introduces a minimal framework that contravenes the notion that effective heuristic design requires more sophisticated hand-engineering pipelines. Potential positive impacts include an improved understanding of how to utilize LLMs effectively in combinatorial optimization (COP), which may benefit applications such as logistics, scheduling, and resource allocation. In addition, by highlighting the trade-off between system complexity and efficiency, this work may encourage the development of simpler and more interpretable algorithm design pipelines, potentially lowering the barrier to heuristic design in less-explored domains. Potential risks stem from the reliance on LLMs, including computational cost and the possibility of unreliable solutions if generated heuristics are applied without validation. In our framework, all candidates are evaluated using deterministic external solvers with feasibility constraints. Human oversight remains important for deployment in real-world settings.

\section{The Use of Large Language Models}
\label{sec:llm_usage}
\paragraph{LLM Usage.}
Large language models (LLMs) are used as a core component of our automated heuristic design (AHD) framework. More precisely, LLMs are used to generate and iteratively refine code implementations of heuristics, guided by structured feedback from external evaluation, with the goal of improving solution quality for combinatorial optimization problems (COPs).

Beyond the AHD process itself, LLMs are used only for minor language polishing of the manuscript (e.g., improving clarity and grammar). They are not used to generate scientific claims, design experimental results, or perform any form of automated analysis. All key ideas, methodological contributions, and experimental findings are developed and verified by the authors.

\section{License}
\label{sec:license}

The licenses and URLs of all baselines, open-source datasets, and cited leaderboard websites are listed in Table \ref{tab:resources}.

\begin{table}[h]
\centering
\footnotesize
\setlength{\tabcolsep}{1.6pt}
\begin{tabular}{lll p{6cm}}
\toprule
Resources & Type & License / Availability & URL \\
\midrule

LKH3 & Code & Available for academic research use & http://webhotel4.ruc.dk/~keld/research/LKH-3/ \\
POMO & Code & Available online & https://github.com/yd-kwon/POMO/tree/master \\
DeepACO & Code & MIT License & https://github.com/henry-yeh/DeepACO \\

\midrule
FunSearch & Code & Apache License & https://github.com/google-deepmind/funsearch \\
EoH & Code & MIT License & https://github.com/FeiLiu36/EoH \\
ReEvo & Code & MIT License & https://github.com/ai4co/reevo \\
MCTS-AHD & Code & MIT License & https://github.com/zz1358m/MCTS-AHD-master \\
\midrule
Artificial Analysis & Leaderboard & Public website & \url{https://artificialanalysis.ai/} \\
\bottomrule
\end{tabular}
\caption{Licenses for codebases, datasets, and public leaderboard for benchmark scores.}
\label{tab:resources}
\end{table}

\newpage
\input{checklist.tex}

\end{document}

%% file: checklist.tex
\section*{NeurIPS Paper Checklist}

\begin{enumerate}

\item {\bf Claims}
    \item[] Question: Do the main claims made in the abstract and introduction accurately reflect the paper's contributions and scope?
    \item[] Answer: \answerYes{} 
    \item[] Justification: All claims in the abstract and introduction—--proposing AHI, ICE, and SimpleEvol, and finding that lower-complexity frameworks yield higher ICE---are directly supported by the experimental results in Section \ref{sec:experiments}.

\item {\bf Limitations}
    \item[] Question: Does the paper discuss the limitations of the work performed by the authors?
    \item[] Answer: \answerYes{} 
    \item[] Justification: Sec. \ref{sec:limitations}

\item {\bf Theory assumptions and proofs}
    \item[] Question: For each theoretical result, does the paper provide the full set of assumptions and a complete (and correct) proof?
    \item[] Answer: \answerNA{} 
    \item[] Justification: The paper makes no theoretical claims requiring formal proofs.

    \item {\bf Experimental result reproducibility}
    \item[] Question: Does the paper fully disclose all the information needed to reproduce the main experimental results of the paper to the extent that it affects the main claims and/or conclusions of the paper (regardless of whether the code and data are provided or not)?
    \item[] Answer: \answerYes{} 
    \item[] Justification: Sec. \ref{sec:intelligence_conversion_model},\ref{sec:simpleEvol_main} and App.~\ref{sec: Prompts_used},\ref{sec:experimental_setups},\ref{sec:problem_descriptions},\ref{sec:calculation_of_metrics},\ref{sec:ice_raw_results}.

\item {\bf Open access to data and code}
    \item[] Question: Does the paper provide open access to the data and code, with sufficient instructions to faithfully reproduce the main experimental results, as described in supplemental material?
    \item[] Answer: \answerNo{} 
    \item[] Justification: The source code is available at https://github.com/HenryZhu1029/SimpleEvol-Master.

\item {\bf Experimental setting/details}
    \item[] Question: Does the paper specify all the training and test details (e.g., data splits, hyperparameters, how they were chosen, type of optimizer) necessary to understand the results?
    \item[] Answer: \answerYes{} 
    \item[] Justification: App. \ref{sec:experimental_setups}. They are also briefly described in Sec. \ref{sec:experiments}.

\item {\bf Experiment statistical significance}
    \item[] Question: Does the paper report error bars suitably and correctly defined or other appropriate information about the statistical significance of the experiments?
    \item[] Answer: \answerYes{} 
    \item[] Justification: We report leave-one-out analysis with means and standard deviations in Appendix~\ref{sec:ICE_variants}, and results are averaged over multiple runs on all our experiments.

\item {\bf Experiments compute resources}
    \item[] Question: For each experiment, does the paper provide sufficient information on the computer resources (type of compute workers, memory, time of execution) needed to reproduce the experiments?
    \item[] Answer: \answerYes{} 
    \item[] Justification: App. \ref{sec:implementation_details}, \ref{sec:cost_analysis}.
    
\item {\bf Code of ethics}
    \item[] Question: Does the research conducted in the paper conform, in every respect, with the NeurIPS Code of Ethics \url{https://neurips.cc/public/EthicsGuidelines}?
    \item[] Answer: \answerYes{} 
    \item[] Justification: This paper studies algorithm design automation and raises no ethical concerns.

\item {\bf Broader impacts}
    \item[] Question: Does the paper discuss both potential positive societal impacts and negative societal impacts of the work performed?
    \item[] Answer: \answerYes{} 
    \item[] Justification: App. \ref{sec:impact_statement}.
    
\item {\bf Safeguards}
    \item[] Question: Does the paper describe safeguards that have been put in place for responsible release of data or models that have a high risk for misuse (e.g., pre-trained language models, image generators, or scraped datasets)?
    \item[] Answer: \answerNA{} 
    \item[] Justification: \answerNA{}

\item {\bf Licenses for existing assets}
    \item[] Question: Are the creators or original owners of assets (e.g., code, data, models), used in the paper, properly credited and are the license and terms of use explicitly mentioned and properly respected?
    \item[] Answer: \answerYes{} 
    \item[] Justification: App. \ref{sec:license}

\item {\bf New assets}
    \item[] Question: Are new assets introduced in the paper well documented and is the documentation provided alongside the assets?
    \item[] Answer: \answerYes{} 
    \item[] Justification: The SimpleEvol code and prompts are documented in Section \ref{sec:simpleEvol_main} and Appendix \ref{sec: Prompts_used}, and will be released upon acceptance.

\item {\bf Crowdsourcing and research with human subjects}
    \item[] Question: For crowdsourcing experiments and research with human subjects, does the paper include the full text of instructions given to participants and screenshots, if applicable, as well as details about compensation (if any)? 
    \item[] Answer: \answerNA{} 
    \item[] Justification: \answerNA{}

\item {\bf Institutional review board (IRB) approvals or equivalent for research with human subjects}
    \item[] Question: Does the paper describe potential risks incurred by study participants, whether such risks were disclosed to the subjects, and whether Institutional Review Board (IRB) approvals (or an equivalent approval/review based on the requirements of your country or institution) were obtained?
    \item[] Answer: \answerNA{} 
    \item[] Justification: \answerNA{}

\item {\bf Declaration of LLM usage}
    \item[] Question: Does the paper describe the usage of LLMs if it is an important, original, or non-standard component of the core methods in this research? Note that if the LLM is used only for writing, editing, or formatting purposes and does \emph{not} impact the core methodology, scientific rigor, or originality of the research, declaration is not required.
    \item[] Answer: \answerYes{} 
    \item[] Justification: App. \ref{sec:llm_usage}

\end{enumerate}